%% file: sample-sigconf.tex
\documentclass[sigconf, screen]{acmart}
\usepackage{multirow}
\usepackage{hhline}
\usepackage{bbding}
\usepackage{colortbl}
\usepackage{xcolor}
\usepackage{balance}

\newcommand{\systemname}{VLMPlanner}
\AtBeginDocument{%
  }

\setcopyright{acmlicensed}
\copyrightyear{2025}
\acmYear{2025}
\acmDOI{XXXXXXX.XXXXXXX}
\acmConference[Conference acronym 'XX]{Make sure to enter the correct
  conference title from your rights confirmation email}{June 03--05,
  2025}{Dublin, Ireland}
\acmISBN{978-1-4503-XXXX-X/2018/06}

\begin{document}

\title{VLMPlanner: Integrating Visual Language Models with Motion Planning}


\author{Zhipeng Tang}
\affiliation{%
  \institution{University of Science and Technology of China}
  \city{Hefei}
  \country{Chine}}
\email{tangzhipeng@mail.ustc.edu.cn}

\author{Sha Zhang}
\authornote{Corresponding author}
\affiliation{%
  \institution{University of Science and Technology of China}
  \city{Hefei}
  \country{China}
}
\email{zhsh1@mail.ustc.edu.cn}

\author{Jiajun Deng}
\affiliation{%
  \institution{University of Adelaide}
  \city{Adelaide}
  \country{Australia}
}
\email{jiajun.deng@adelaide.edu.cn}

\author{Chenjie Wang}
\affiliation{%
  \institution{	Institute of Artificial Intelligence, Hefei Comprehensive National Science Center}
  \city{Hefei}
  \country{Chine}}
\email{wangchenjie@iai.ustc.edu.cn}

\author{Guoliang You}
\affiliation{%
  \institution{University of Science and Technology of China}
  \city{Hefei}
  \country{Chine}}
\email{glyou@mail.ustc.edu.cn}

\author{Yuting Huang}
\affiliation{%
  \institution{University of Science and Technology of China}
  \city{Hefei}
  \country{Chine}}
\email{yutinghuang@mail.ustc.edu.cn}

\author{Xinrui Lin}
\affiliation{%
  \institution{University of Science and Technology of China}
  \city{Hefei}
  \country{Chine}}
\email{linxinrui@mail.ustc.edu.cn}

\author{Yanyong Zhang}
\authornotemark[1]
\affiliation{%
  \institution{University of Science and Technology of China}
  \city{Hefei}
  \country{China}
}
\email{yanyongz@ustc.edu.cn}

\renewcommand{\shortauthors}{Tang et al.}

\begin{abstract}
\input{Section/0-abstract}
\end{abstract}





\maketitle

\input{Section/1-intro}

\input{Section/2-related}

\input{Section/3-method}

\input{Section/4-exp}
\input{Section/5-conclusion}



\clearpage
\bibliographystyle{ACM-Reference-Format}
\balance
\bibliography{sample-base}


%

\end{document}

%% file: Section/0-abstract.tex
Integrating large language models (LLMs) into autonomous driving motion planning has recently emerged as a promising direction, offering enhanced interpretability, better controllability, and improved generalization in rare and long-tail scenarios.
However, existing methods often rely on abstracted perception or map-based inputs, missing crucial visual context, such as fine-grained road cues, accident aftermath, or unexpected obstacles, which are essential for robust decision-making in complex driving environments.
To bridge this gap, we propose VLMPlanner, a hybrid framework that combines a learning-based real-time planner with a vision-language model (VLM) capable of reasoning over raw images.
The VLM processes multi-view images to capture rich, detailed visual information and leverages its common-sense reasoning capabilities to guide the real-time planner in generating robust and safe trajectories. 
Furthermore, we develop the Context-Adaptive Inference Gate (CAI-Gate) mechanism that enables the VLM to mimic human driving behavior by dynamically adjusting its inference frequency based on scene complexity, thereby achieving an optimal balance between planning performance and computational efficiency.
We evaluate our approach on the large-scale, challenging nuPlan benchmark, with comprehensive experimental results demonstrating superior planning performance in scenarios with intricate road conditions and dynamic elements. 
Code will be available. 

%% file: Section/1-intro.tex
\section{Introduction}
\begin{figure}[th]
    \centering
    \includegraphics[width=1.0\columnwidth]{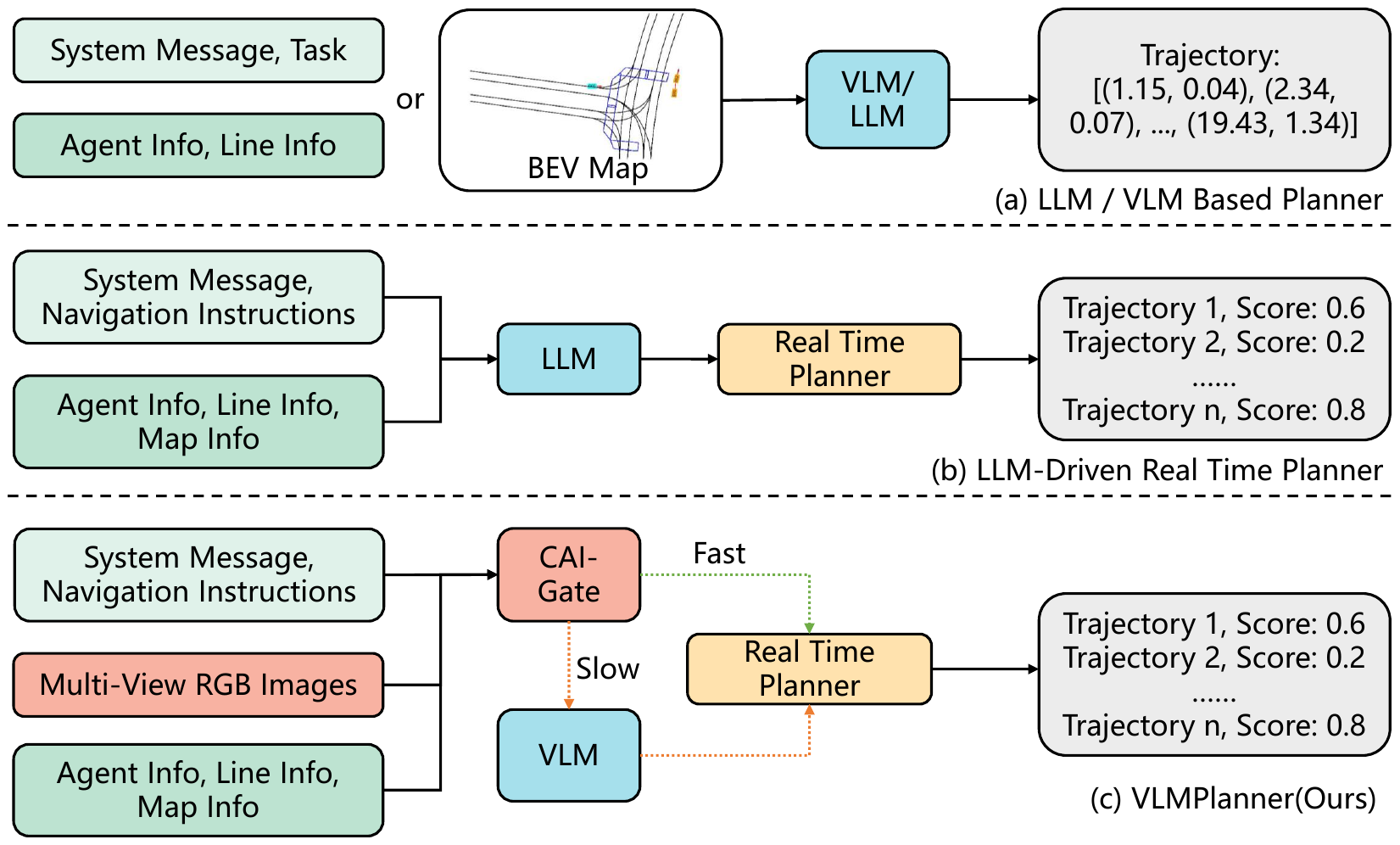}
    \caption{Comparison of LLM-related planner methods. (a) LLM/VLM based Planner: transforms driving scene data into structured natural language representations or BEV map for trajectory planning. (b) LLM-Driven Real Time Planner: integrates LLMs with real-time planner. (c) Our VLMPlanner: Take Multi-View RGB images as supplementary input and develop Context-Adaptive Inference Gate (CAI-Gate) to dynamically adjust VLM's inference frequency.}
    \label{fig:improve}
\end{figure}

Motion planning is a core component in autonomous driving systems~\cite{zhang2020autosys, nuplan}, responsible for generating safe, comfortable, and efficient trajectories for the ego vehicle. Despite notable progress, traditional approaches~\cite{hallgarten2023prediction, huang2023gameformer, renz2022plant, cheng2024rethinking, cheng2024pluto, wang2024lhpf, yang2024diffusion, huang2024dtpp, hu2024solving, huang2023differentiable, hu2023imitation} often struggle in complex, dynamic scenarios, especially when facing long-tail corner cases. 
In parallel, recent advances in large language models (LLMs)~\cite{grattafiori2024llama, achiam2023gpt4} have shown strong potential to address these challenges, demonstrating impressive capabilities in commonsense reasoning and cross-domain knowledge transfer. With further instruction tuning, they can develop a foundational understanding of traffic rules, driving scenarios, and behavioral patterns, enabling more informed and context-aware planning in complex driving environments. These advantages position LLM-based motion planning as a promising and attractive research direction.



As illustrated in Fig~\ref{fig:improve}(a), prior LLM-based planning methods~\cite{cui2024survey, yang2023llm4drive, sharan2023llm, mao2023gpt, mao2023language, sha2023languagempc, shao2024lmdrive} have leveraged the reasoning capabilities of LLMs/VLMs by converting driving scene data into structured natural language representations or BEV maps~\cite{zheng2024planagent, yao2024calmm} for trajectory planning. 
However, these approaches are inherently constrained by token limitations, which restrict the ability of LLMs to accurately represent fine-grained features in complex scenarios. 
Such limitations can lead to ambiguities or the loss of critical environmental details essential for safe navigation. 
Moreover, the reliance on language outputs for trajectory generation exacerbates the models' deficiencies in numerical computation and geometric reasoning, ultimately compromising the accuracy of the trajectories.

To address these challenges, several approaches~\cite{chen2024asynchronous, sharan2023llm} advocate integrating LLMs into the real-time planner rather than replacing it, thereby circumventing the need for direct trajectory generation by LLMs. 
This integrated design not only improves prediction accuracy but also enables finer trajectory control, as demonstrated in Fig~\ref{fig:improve}(b). 
Moreover, AsyncDrive~\cite{chen2024asynchronous} addresses token limitations by pre-encoding perception and map information prior to feeding it into the LLM.
However, such approaches typically restrict their inputs to perception outputs, limiting the effective utilization of LLM/VLM capabilities—capabilities that are ideally suited for addressing long-tail scenarios, emergency events, and traffic commonsense. 
While perception results can support basic traffic logic, they often compress information, resulting in a significant loss of environmental detail. 
For instance, traditional perception and map data offer limited insight into complex, rare scenes, failing to predict potential hazards such as large road puddles, bicycles executing signaling turns, and tidal lanes. 
These critical cues might not be explicitly represented in perception data or map data or would require complex inference to extract, whereas image data can directly deliver such nuanced details.

Motivated by these observations, we propose a novel hybrid planning framework—\systemname—that integrates a conventional real-time planner with an enhanced VLM-based module that accepts multi-view images as input. 
As depicted in Fig~\ref{fig:improve}(c), our approach leverages multi-view image data to capture fine-grained environmental details, thereby overcoming the limitations of relying solely on compressed perception or map outputs.
Furthermore, the rich details provided by multi-view images enable the analysis of emergent and complex scenarios, thereby 
pinpointing when VLM inference is required.
To achieve this, we design the Context-Adaptive Inference Gate (CAI-Gate), which dynamically modulates the VLM’s involvement based on scene complexity, ensuring optimal utilization of its reasoning power while maintaining computational efficiency and planning accuracy.

Within our framework, the VLM-based module extracts rich semantic cues—such as subtle changes in traffic flow, nuanced vehicle maneuvers, and early indicators of pedestrian intent—that are subsequently fused with standard perception outputs to inform trajectory planning. To address token limitations, we employ 3D positional encoding~\cite{liu2022petr} and a 3D-aware module similar to Q-former architecture~\cite{li2022blip} to aggregate multi-view image features. This design not only substantially reduces the number of tokens required by the VLM but also enhances 3D comprehension by elevating visual features into the 3D space.
Furthermore, to improve the domain adaptability of vision language models for autonomous driving and establish scene-relevant commonsense knowledge, we have developed two specialized fine-tuning datasets—DriveVQA and ReasoningVQA—derived from the nuPlan dataset with support from GPT-4’s \cite{achiam2023gpt4} generative capabilities. DriveVQA focuses on high-level driving instructions, control commands, and waypoint-related queries, while ReasoningVQA guides large models to analyze trajectories comprehensively based on surrounding scene information and traffic regulations, providing detailed rationales for the corresponding decisions.

To evaluate our framework, we adopt the nuPlan \cite{nuplan} and curate a test suite targeting challenging long-tail scenarios in both open-loop and closed-loop configurations. We adapt the evaluation protocol to accommodate image inputs, and the experiments demonstrate that our planning framework outperforms state-of-the-art methods. Furthermore, ablation studies confirm that even with reduced VLM inference frequency, the model can maintain robust performance.
In summary, our contributions are as follows:
\begin{itemize}
    \item We integrate the VLM into the autonomous motion planning framework by leveraging high-fidelity multi-view image data to capture rich and fine-grained environmental details. This integration enables our system to extract subtle visual cues—critical for addressing corner cases and complex scenarios—thereby enhancing the planner’s accuracy and robustness in challenging driving conditions.
    \item We construct two comprehensive image-text datasets — DriveVQA and ReasoningVQA — specifically tailored for autonomous driving scene understanding.  
    \item We propose an adaptive motion planner that dynamically adjusts LLM inference frequency based on scene complexity, effectively addressing real-time computational challenges in large-scale planning models. 
    \item Our planning framework outperforms state-of-the-art methods in both open-loop and closed-loop settings, and ablation studies confirm that even with a reduced VLM inference frequency, the model maintains robust performance.
\end{itemize}

%% file: Section/2-related.tex
\section{Related Work}

\subsection{End-to-end Autonomous Driving}
End-to-end (e2e) methods \cite{zeng2019end, hu2022st, hu2023planning} directly predict future trajectories from raw sensor inputs. In recent years, these end-to-end approaches have evolved from their initial image-only architectures \cite{codevilla2019exploring} to progressively incorporate multimodal fusion frameworks \cite{chitta2022transfuser, jia2023think}, and have further advanced toward modularized end-to-end structures that integrate perception, prediction, and planning into a unified model \cite{hu2022st, jiang2023vad}. Despite their significant developmental potential, most end-to-end methods still face critical challenges regarding the authenticity and diversity of simulated agents during testing phases, primarily due to the substantial sim-to-real gap between virtual environments and the physical world. 

\subsection{Learning-Based Motion Planning for Autonomous Driving}
The modular architecture for autonomous driving systems typically comprises three sequential components: perception, prediction, and planning. Within this architecture, motion planning predicts future trajectories with the aim of balancing driving safety, comfort, efficiency, and progress along the intended route. 
By defining well-structured data interfaces between modules enables focused research and optimization of individual tasks.

Current research in autonomous driving planners primarily focuses on learning-based approaches, which can be broadly categorized into imitation learning methods \cite{hallgarten2023prediction, huang2023gameformer, renz2022plant, cheng2024rethinking, cheng2024pluto, wang2024lhpf, yang2024diffusion, huang2024dtpp, hu2024solving, huang2023differentiable, hu2023imitation} and offline reinforcement learning methods \cite{kendall2019learning, li2024boosting, scheel2022urban}. These approaches aim to replicate human expert driving behaviors by learning from large-scale real-world driving datasets. GameFormer \cite{huang2023gameformer} employs a hierarchical trajectory prediction framework based on DETR \cite{zhu2020deformable} architecture. DTPP \cite{huang2024dtpp} introduces a differentiable framework for the joint training of ego-conditioned prediction and cost models. PlanTF \cite{cheng2024rethinking} introduces an attention-based state dropout encoder and augmentation techniques.  However, they face limitations due to the scope of the datasets and model complexity, leaving substantial room for improvement in areas like routing information comprehension and environmental awareness.

\subsection{Language-Augmented Motion Planning with LLMs}
The rapid advancement of large language models (LLMs) has garnered significant attention. Models such as GPT-4 \cite{achiam2023gpt} and Llama3 \cite{grattafiori2024llama} have been trained on extensive textual data, demonstrating remarkable generalization and reasoning capabilities. An increasing number of studies \cite{chen2023towards, chen2024driving, cui2024receive, fu2024drive, han2024dme, jin2023surrealdriver, liu2023mtd, ma2024dolphins, ma2024lampilot, mao2023gpt, mao2023language, nie2024reason2drive, renz2022plant, sha2023languagempc, sima2024drivelm, wang2023chatgpt, wang2023drivemlm, wang2023empowering, wen2023dilu, xu2024drivegpt4, yuan2024rag, xu2024vlm, li2025generative, hwang2024emma, jiang2024senna, ding2024holistic, wang2024omnidrive} have begun to explore the potential of leveraging LLMs’ decision-making abilities in the domain of autonomous driving planning. 

Some studies \cite{zheng2024planagent, mao2023gpt, mao2023language, sharan2023llm, yao2024calmm} convert scene information (including the state of the ego vehicle as well as information about obstacles, pedestrians, and other vehicles) into linguistic information or visualized BEV maps, which are subsequently fed into LLMs/VLMs for scene comprehension and trajectory planning . 
PlanAgent \cite{zheng2024planagent} employs BEV maps to facilitate scene understanding and provide guidance for the underlying planner's trajectory planning. 
LLM-ASSIST \cite{sharan2023llm} proposes a hybrid planner that integrates LLMs with rule-based methods to improve autonomous driving. 
To address context length constraints and achieve more precise information encoding, AsyncDriver \cite{chen2024asynchronous} operates by inputting encoded map information and navigation directives into an LLM to guide the underlying planner's trajectory generation. Furthermore, it implements a decoupling mechanism between the LLM and real-time planner, enabling the LLM to execute inferences at a fixed frequency lower than that of the real-time planner. This architectural innovation partially mitigates the characteristic high latency and substantial computational overhead associated with contemporary LLM-based planners. LLM-Assist \cite{sharan2023llm} similarly decouples the LLM from the rule-based planner, with the LLM intervening only when the planned trajectory fails to meet predefined scoring thresholds. 
However, these approaches fail to capture certain critical contextual details during scene understanding. To address this limitation, we introduce multi-view images and employ VLMs to comprehend detailed driving scene information. Furthermore, we propose a CAI-Gate mechanism that dynamically adapts the VLM's inference frequency, thereby achieving an optimal balance between planning performance and computational efficiency.


%% file: Section/3-method.tex
\section{Methodology}

\begin{figure*}[th]
    \centering
    \resizebox{\textwidth}{!}{
        \includegraphics[width=\textwidth]{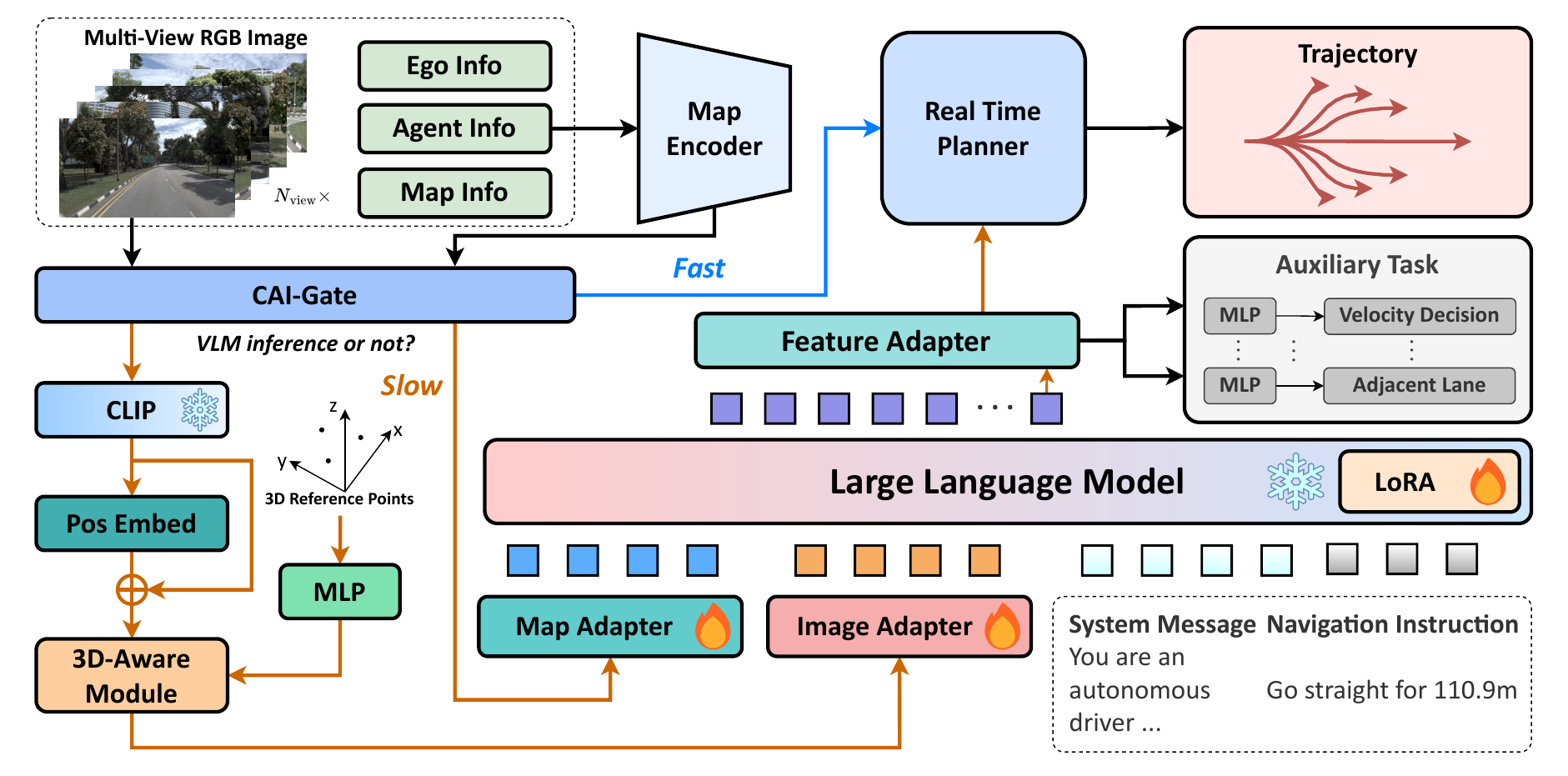}
    }
    \caption{The overall framework of \textbf{VLMPlanner}. Our method comprises two key components: (1) the real-time planner, which processes solely map information to generate rapid trajectory planning (denoted by blue lines in the diagram), and (2) the VLM guidance, which performs scene understanding by integrating real-time multi-view RGB images with map information to direct the execution of planning decisions (represented by black lines in the illustration).}
    \label{fig:overview}
\end{figure*}

\subsection{Overall Framework}
\label{sec:overall}

As illustrated in Fig. \ref{fig:overview}, we present \systemname, a hybrid planning framework that integrates a learning-based real-time planner with VLM. The real-time planner exclusively processes map information as input, which undergoes encoding and decoding operations to generate predicted future trajectories. The VLM receives multi-view images, map information, system message, and navigation instruction as inputs. These heterogeneous data streams are individually encoded and subsequently integrated as composite inputs to the VLM. The VLM then utilizes its output hidden states to provide planning guidance to the real-time planner. The CAI-Gate module, which takes multi-view images and map information as inputs, evaluates scene complexity to dynamically adjust the VLM's inference frequency during the inference phase, thereby achieving an optimal balance between performance and computational efficiency.

In the following sections, we first present a detailed exposition of the model architecture design in Sec. \ref{sec:model_arch}, followed by the implementation specifics of the CAI-Gate module in Sec. \ref{sec:cai-gate}. Subsequently, Sec. \ref{sec:data_generation} elaborates on the composition and construction pipeline of  DriveVQA and ReasoningVQA. Finally, Sec. \ref{sec:training_detail} delineates the procedural workflow of our two-stage training paradigm.

\subsection{Model Architecture}
\label{sec:model_arch}
\textbf{MultiModal Input.} As illustrated in Fig. ~\ref{fig:overview}, our model input comprises four distinct components: (1) system message, (2) navigation instructions, (3) encoded map features, and (4) encoded image features. These inputs are structurally organized as follows:
\begin{equation}
    \label{eq:prompt}
    \begin{aligned}
        & \underbrace{\text{You are an autonomous driver ...}}_{\text{system message}} \ \ \ \ 
        \underbrace{F_{\text{map}}^{(1)}, \dots, F_{\text{map}}^{(M)}}_{\substack{\text{map tokens}}} \\
        & \underbrace{F_{\text{img}}^{(1)}, \dots, F_{\text{img}}^{(N)}}_{\text{image tokens}} \ \ \ \  \underbrace{\text{Go straight for 110.9m}}_{\text{navigation instruction}}
    \end{aligned}
\end{equation}
where $F_{\text{map}} \in \mathbb{R}^{M\times D_{\text{llm}}}$ denotes the map information tokens while $F_{\text{img}} \in \mathbb{R}^{N\times D_{\text{llm}}}$ represents the multi-view image tokens. The complete prompt information can be found in the Appendix.

\noindent \textbf{Map Information.} For map information processing, we extract the following data from the dataset: the ego vehicle's hitorical state information $S_\text{ego} \in \mathbb{R}^{1\times T_h \times d_e}$, historical state information $S_{\text{neighbor}} \in \mathbb{R}^{N_n \times T_h \times d_n}$ of neighbor agents where $d_e, d_n$ denote the number of the state attributes for ego vehicle and neighboring agents, respectively, while $T_h$ represents the number of historical time steps, and surrounding map information including lane features $M_{\text{lane}} \in \mathbb{R}^{N_l \times N_p \times d_p}$ and crosswalk features $M_{\text{crosswalk}} \in \mathbb{R}^{N_c \times N_p \times d_p}$. Specifically, we identify the $N_l$ nearest lanes and $N_c$ nearest crosswalks relative to the ego vehicle, each containing $N_p$ points with $d_p$ attributes. All information is transformed into the ego-vehicle-centric local coordinate frame, and any missing positions in the tensors are padded with zeros. We employ the MapEncoder from Gameformer \cite{huang2023gameformer} to encode the aforementioned data, followed by MLPAdapter, a MLP layer, to match the dimensions of text token embeddings, and obtain the map features $F_\text{map} \in \mathbb{R}^{M \times D_{\text{llm}}}$. 
\begin{equation}
    \tilde{F}_{\text{map}} = \text{MapEncoder} (S_{\text{ego}}, S_{\text{neighbor}}, M_{\text{lane}}, M_{\text{crosswalk}}) \\
\end{equation}
\begin{equation}
    F_{\text{map}} = \text{MapAdapter}(\tilde{F}_{\text{map}})
\end{equation}

\noindent \textbf{Multi-View Images.} For visual information processing, we extract multi-view images $I \in \mathbb{R}^{N_\text{view}\times 3 \times H\times W}$ at the current timestep and employ CLIP \cite{radford2021learning} to directly encode the images, obtaining visual features $\tilde{F}_{\text{img}} \in \mathbb{R}^{N_{\text{view}}\times HW\times C}$. To incorporate spatial awareness, we feed both the image features and their corresponding positional encodings $P_\text{3D}\in \mathbb{R}^{N_\text{view}\times HW \times C}$ into 3D-aware module similar to Q-Former \cite{li2022blip}. In 3D-aware module, we initialize a set of learnable reference points uniformly distributed in [0,1] range, where 3D coordinates are generated for each point via a compact MLP network to produce query $Q_{\text{3D}} \in \mathbb{R}^{N \times C}$. Through cross-attention mechanisms, we project the image features into 3D space while simultaneously reducing the number of image tokens input to the VLM. Finally, we utilize an ImgAdapter to align the feature dimensions with the VLM's requirements, yielding the processed image features $F_{\text{img}} \in \mathbb{R}^{N\times D_{\text{llm}}}$.
\begin{equation}
    \tilde{F}_{\text{img}} = \text{3D-Aware} \big({\text{CLIP}} (I), P_{\text{3D}}, Q_{\text{3D}} \big)
\end{equation}
\begin{equation}
    F_{\text{img}} = \text{ImgAdapter} ( \tilde{F}_{\text{img}} )
\end{equation}

\noindent \textbf{Navigation Instruction.} During the training phase, our navigation instructions are directly derived from the ground truth of the ego vehicle's future trajectory, where the actual future path is converted into textual descriptions (e.g., "Go straight for 110.9m"). During inference, the navigation instructions are generated based on lane information and navigation data from the map. Please refer to the Appendix for detailed implementation. 

\noindent \textbf{VLM-Guided Real-Time Planner.} We input the map embeddings, image embeddings, and language embeddings into the VLM and obtain the final hidden layer features $H = \{h_i\}_{i=1}^n$. Subsequently, we employ the Adaptive Injection Block from AsyncDriver \cite{chen2024asynchronous} to inject the scene information comprehended by the VLM into the real-time planner, thereby generating the final planning results. 
Specifically, we first extract the hidden feature of the last token $h_{n}$ and project it through the feature adapter.
\begin{equation}
    \tilde{h} = \text{FeatureAdapter}(h_{n})
\end{equation}
Then, we extend the decoding layer of the real-time planner:
\begin{equation}
    s^{l+1} = g \cdot \text{MultiHeadAttn}(Q,K_{\tilde{h}}, V_{\tilde{h}}) + \text{MultiHeadAttn}(Q,K_{s^l}, V_{s^l})
\end{equation}
where $g$ is the value of adaptive gate, $Q$ denotes query in original decoder layer, $K_i$ and $V_i$ represent key and value respectively of feature $i$, and $s^l$ note the scene feature of the $l$-th decoder layer. 

\begin{table*}[t]
    \caption{Evaluation on nuPlan Open-Loop Challenges on Open-Hard20 split. The best result are highlighted in bold. Type 0-13 represent the 14 official scenario types of the nuPlan challenge 2023 \cite{nuplan}. "Ours(GameFormer)" and "Ours(PlanTF)" denote our framework's implementations utilizing GameFormer and PlanTF as the real-time planners, respectively.}
    \label{tab:open_result}
    \resizebox{\textwidth}{!}{
        \begin{tabular}{cc*{14}{c}}
            \toprule
            Methods & \cellcolor{gray!25} score & type0 & type1 & type2 & type3 & type4 & type5 & type6 & type7 & type8 & type9 & type10 & type11 & type12 & type13 \\
            \midrule
            PDM-Hybrid \cite{dauner2023parting}
            & \cellcolor{gray!25} $58.21$ 
            & $64.48$ & $47.52$ & $45.89$ & $85.07$ & $36.51$ & $11.41$ & $63.63$ 
            & $53.15$ & $81.38$ & $88.07$ & $79.15$ & $33.14$ & $63.66$ & $85.82$ 
            \\
            PLUTO \cite{cheng2024pluto}
            & \cellcolor{gray!25} $57.21$ 
            & $38.80$ & $59.34$ & $34.77$ & $70.61$ & $43.48$ & $\textbf{65.46}$ & $42.88$ 
            & $46.69$ & $69.53$ & $84.40$ & $69.22$ & $46.84$ & $74.54$ & $83.08$ 
            \\
            PlanTF \cite{cheng2024rethinking}
            & \cellcolor{gray!25} $78.86$ 
            & $56.16$ & $\textbf{79.97}$ & $66.40$ & $\mathbf{91.20}$ & $69.08$ & $62.85$ & $77.95$ 
            & $\mathbf{84.50}$ & $\mathbf{85.33}$ & $\mathbf{92.30}$ & $93.41$ & $80.71$ & $89.80$ & $87.37$ 
            \\
            DTPP \cite{huang2024dtpp}
            & \cellcolor{gray!25} $59.71$ 
            & $44.00$ & $43.51$ & $28.98$ & $73.96$ & $59.19$ & $25.55$ & $47.26$ 
            & $69.95$ & $83.15$ & $88.46$ & $76.37$ & $62.10$ & $78.60$ & $85.44$
            \\
            GameFormer \cite{huang2023gameformer}
            & \cellcolor{gray!25} $76.22$ 
            & $76.95$ & $67.63$ & $55.17$ & $87.01$ & $\mathbf{72.71}$ & $59.62$ & $\mathbf{80.42}$ 
            & $82.02$ & $75.85$ & $86.08$ & $84.70$ & $78.91$ & $85.86$ & $\mathbf{91.19}$ 
            \\
            AsyncDriver \cite{chen2024asynchronous}
            & \cellcolor{gray!25} $75.24$ 
            & $\mathbf{78.04}$ & $63.36$ & $59.40$ & $86.30$ & $66.46$ & $64.05$ & $68.54$ 
            & $77.53$ & $81.50$ & $89.21$ & $81.87$ & $82.60$ & $75.90$ & $90.46$ 
            \\
            \textbf{Ours (GameFormer)}
            & \cellcolor{gray!25} $76.54$
            & $76.64$ & $71.01$ & $64.48$ & $86.86$ & $61.33$ & $60.42$ & $79.30$ 
            & $77.61$ & $77.48$ & $86.51$ & $86.35$ & $\mathbf{82.78}$ & $86.64$ & $91.13$ 
            \\
            \textbf{Ours (PlanTF)}
            & \cellcolor{gray!25} $\mathbf{79.12}$ 
            & $56.06$ & $79.88$ & $\mathbf{69.44}$ & $91.18$ & $72.06$ & $63.08$ & $78.81$ 
            & $80.55$ & $85.10$ & $92.08$ & $\mathbf{93.82}$ & $80.94$ & $\mathbf{90.00}$ & $87.20$ 
            \\
            \bottomrule
        \end{tabular}
    }
\end{table*}

\subsection{Context-Adaptive Inference Gate}
\label{sec:cai-gate}
Previous LLM-based planning algorithms typically required the LLM to participate in every inference cycle. However, this approach proves inherently unsuitable for latency-sensitive motion planning tasks. Inspired by human driving behavior, where drivers dynamically allocate attention based on scenario complexity \cite{kircher2017minimum, liu2021drivers}, we decouple the VLM from the real-time planner and design Context-Adaptive Inference
Gate (CAI-Gate) that dynamically adjusts the VLM's inference frequency according to scene complexity, thereby blance planning performance with computational efficiency.

\noindent \textbf{Learning-Based CAI-Gate.} We employ real-time multi-view image inputs and utilize a compact EfficientNet-B0 \cite{tan2019efficientnet} based network for rapid assessment of driving scene complexity. Specifically, we first leverage multi-view scene images paired with linguistic prompts to classify scene complexity levels (categorized into $5$ grades) using Gemini, thereby obtaining a dataset of 7,503 images-ground truth pairs, with $70\%$ allocated for training and $30\%$ for testing. Subsequently, we train an ordinal regression model to classify scene complexity levels, which then determines the appropriate inference frequency for the VLM.

\noindent \textbf{Rule-Based CAI-Gate.} Additionally, we have developed a rule-based methodology for classifying driving scene complexity. The complexity of driving scenarios is determined by several key factors: the number and proximity of surrounding vehicles and pedestrians, the quantity of lanes at the ego vehicle's position, the ego vehicle's current speed and acceleration, and its distance to intersections or turns. By individually assessing and aggregating these metrics, we ultimately derive a comprehensive scene complexity rating. More detailed specifications can be found in the Appendix. 

\subsection{Data Generation}
\label{sec:data_generation}
Existing VLMs possess extensive commonsense knowledge; however, they exhibit limited familiarity with autonomous driving scenarios, and currently suffer from a scarcity of image-text data specifically tailored for such contexts. To enhance VLMs' comprehension of autonomous driving scenes and instructions, we have constructed two multi-view image-text datasets based on nuPlan \cite{caesar2021nuplan}: DriveVQA and ReasoningVQA. 

\noindent \textbf{DriveVQA.} is specifically designed to enhance VLMs' comprehension of autonomous driving instructions, trajectories, and control signals. Our methodology involves: (1) extracting the ego vehicle's state information, historical/future trajectories, and multi-view images from the dataset; (2) algorithmically generating corresponding textual descriptions; and (3) systematically combining autonomous driving instructions, trajectory data, and control signals through random pairwise permutations, which are then integrated with the multi-view images to form comprehensive visual-textual QA pairs.

\noindent \textbf{ReasoningVQA} is specifically designed to enhance VLMs' scene comprehension and reasoning capabilities in driving scenarios. This dataset is constructed through a hybrid approach combining manually designed rules with GPT-4 assisted generation. Our methodology involves: (1) converting raw map data from the original dataset into comprehensive textual descriptions, including precise positional data, historical trajectory information of both ego and surrounding vehicles, and detailed lane specifications; (2) integrating all relevant information; and (3) prompting GPT-4 to generate contextually appropriate planning decisions accompanied by detailed reasoning processes, strictly adhering to traffic regulations and the provided map information.

Our final dataset comprises $49,673$ samples in DriveVQA and $1,099$ samples in ReasoningVQA. Complete data construction details are provided in the Appendix.

\begin{table*}
    \caption{Evaluation results on Close-Hard20 split with nonreactive agents. The best results are highlighted in bold. \textit{Score}: final score in average. \textit{Drivable}: drivable area compliance. \textit{Direct}: driving direction comliance. \textit{Comf}: ego is comfortable. \textit{Prog}: ego progress along expert. \textit{Coll}: no ego at-fault collisions. \textit{Lim}: speed limit compilance. \textit{TTC}: time to collision with bound. "Ours(GameFormer)" and "Ours(PlanTF)" denote our framework's implementations utilizing GameFormer and PlanTF as the real-time planners, respectively.}
    \label{tab:close_nr_result}
        \begin{tabular}{c*{8}{c}}
            \toprule
            Methods    & \cellcolor{gray!25} Score & Drivable & Direct. & Comf. & Prog. & Coll. & Lim. & TTC \\
            \midrule
            PDM-Closed \cite{dauner2023parting}   
            & \cellcolor{gray!25} $33.84$ 
            & $86.56$ & $97.83$ & $79.45$ & $90.96$ & $47.23$ & $99.43$ & $36.36$ \\
            LLM-Assist \cite{sharan2023llm}
            & \cellcolor{gray!25} $31.79$ 
            & $86.56$ & $97.83$ & $61.66$ & $91.76$ & $49.60$ & $\mathbf{99.47}$ & $32.81$ \\
            PLUTO \cite{cheng2024pluto}
            & \cellcolor{gray!25} $58.67$ 
            & $91.30$ & $97.43$ & $83.00$ & $89.73$ & $68.97$ & $98.97$ & $64.82$ \\
            DTPP \cite{huang2024dtpp}
            & \cellcolor{gray!25} $75.09$ 
            & $\mathbf{96.84}$ & $\mathbf{99.60}$ & $94.47$ & $84.03$ & $77.47$ & $99.25$ & $77.47$ \\
            PlanTF \cite{cheng2024rethinking}
            & \cellcolor{gray!25} $74.89$ 
            & $91.70$ & $97.04$ & $\mathbf{95.26}$ & $88.19$ & $89.13$ & $99.41$ & $78.26$ \\
            GameFormer \cite{huang2023gameformer}
            & \cellcolor{gray!25} $73.99$ 
            & $91.70$ & $99.01$ & $93.68$ & $90.27$ & $85.18$ & $99.40$ & $76.28$ \\
            AsyncDriver \cite{chen2024asynchronous}
            & \cellcolor{gray!25} $52.75$ 
            & $94.47$ & $98.81$ & $90.51$ & $\mathbf{92.62}$ & $60.67$ & $99.33$ & $50.99$ \\
            \textbf{Ours (PlanTF)}   
            & \cellcolor{gray!25} $75.82$ 
            & $90.44$ & $98.21$ & $94.42$ & $87.75$ & $89.24$ & $99.45$ & $\mathbf{81.27}$ \\
            \textbf{Ours (GameFormer)}   
            & \cellcolor{gray!25} $\mathbf{76.35}$ 
            & $94.86$ & $99.01$ & $93.28$ & $84.90$ & $\mathbf{89.33}$ & $99.42$ & $79.05$ \\
            \bottomrule
        \end{tabular}
\end{table*}

\begin{table*}
    \caption{Evaluation results on Close-Hard20 split with reactive agents. The best results are highlighted in bold.}
    \label{tab:close_r_result}
        \begin{tabular}{c*{8}{c}}
            \toprule
            Methods    & \cellcolor{gray!25} Score & Drivable & Direct. & Comf. & Prog. & Coll. & Lim. & TTC \\
            \midrule
            PDM-Closed \cite{dauner2023parting}
            & \cellcolor{gray!25} $42.16$ 
            & $90.51$ & $98.81$ & $72.33$ & $85.90$ & $50.20$ & $99.40$ & $41.11$ \\
            LLM-Assist \cite{sharan2023llm}
            & \cellcolor{gray!25} $40.51$ 
            & $88.14$ & $98.02$ & $60.47$ & $87.21$ & $52.57$ & $\mathbf{99.45}$ & $43.48$ \\
            PLUTO \cite{cheng2024pluto}
            & \cellcolor{gray!25} $48.87$ 
            & $90.12$ & $97.04$ & $73.12$ & $79.29$ & $67.58$ & $99.10$ & $60.08$ \\
            DTPP \cite{huang2024dtpp}
            & \cellcolor{gray!25} $61.29$ 
            & $\mathbf{94.86}$ & $\mathbf{99.80}$ & $90.51$ & $83.34$ & $64.82$ & $99.22$ & $64.82$ \\
            PlanTF \cite{cheng2024rethinking}
            & \cellcolor{gray!25} $61.57$ 
            & $92.89$ & $97.83$ & $94.47$ & $91.02$ & $71.34$ & $99.28$ & $63.64$ \\
            GameFormer \cite{huang2023gameformer}
            & \cellcolor{gray!25} $64.51$ 
            & $90.51$ & $96.64$ & $85.38$ & $89.83$ & $72.53$ & $99.06$ & $66.40$ \\
            AsyncDriver \cite{chen2024asynchronous}
            & \cellcolor{gray!25} $54.26$ 
            & $92.09$ & $96.05$ & $82.21$ & $\mathbf{91.56}$ & $61.26$ & $99.08$ & $52.57$ \\
            \textbf{Ours (PlanTF)}
            & \cellcolor{gray!25} $62.02$ 
            & $91.63$ & $98.61$ & $\mathbf{95.62}$ & $90.73$ & $71.31$ & $99.34$ & $64.54$ \\
            \textbf{Ours (GameFormer)}
            & \cellcolor{gray!25} $\mathbf{66.66}$ 
            & $90.91$ & $95.65$ & $84.58$ & $84.29$ & $\mathbf{76.88}$ & $99.37$ & $\mathbf{70.36}$ \\
            \bottomrule
        \end{tabular}
\end{table*}

\begin{figure*}[t]
    \centering
    \resizebox{!}{0.7\textheight}{
        \includegraphics[]{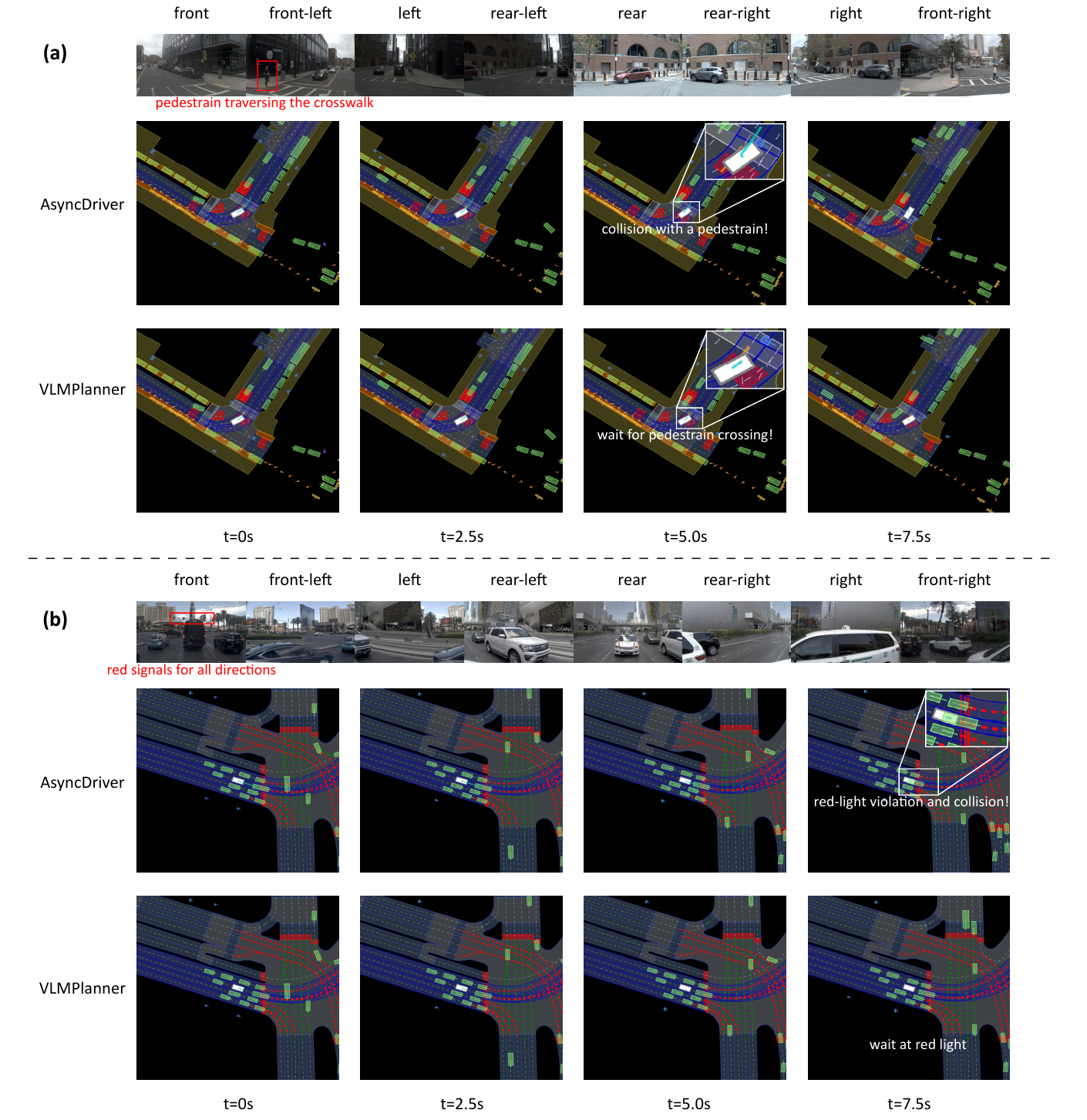}
    }
    \caption{Comparative visualization analysis between our VLMPlanner and AsyncDriver in various driving scenarios. In the map, the white agent denotes the ego vehicle, while green and blue markers correspond to other vehicles and pedestrians, respectively.}
    \label{fig:visualization}
\end{figure*}

\subsection{Training Objectives}
\label{sec:training_detail}
Our training pipeline consists of two distinct phases: a pre-training stage and a fine-tuning stage. During the pretraining phase, we utilize multi-view image-text paired data to enhance the VLM's adaptation capability in autonomous driving scenarios. The finetuning phase exclusively employs $10,000$ sampled data instances from the nuPlan dataset to augment the real-time planner's performance through VLM-based enhancement.

\noindent \textbf{Pre-training Stage.} We conduct two rounds of VLM pretraining using DriveVQA and ReasoningVQA respectively. The first round uses only DriveVQA data to enhance the VLM's understanding of autonomous driving instructions, trajectories, and control signals. The second round uses ReasoningVQA data with a small amount of DriveVQA data to improve the VLM's understanding of autonomous driving scenarios and traffic rules, as well as its reasoning ability. Both rounds employ LoRA for training.

\noindent \textbf{Fine-tuning Stage.} 
We commence by delineating the composition of our training data. For the fine-tuning stage, the training samples are exclusively drawn from the sensor-equipped subset of the nuPlan dataset. We have constructed a curated fine-tuning dataset comprising precisely 10,000 instances, while maintaining identical category distribution with the complete nuPlan trainval set.
Following the GameFormer \cite{huang2023gameformer} paradigm, each data frame incorporates vectorized scene representations containing: (1) $20$-frame historical trajectories for both the ego vehicle and a maximum of $20$ surrounding agents, (2) global map data encompassing the ego vehicle's vicinity, and (3) $80$-frame future trajectories of the ego vehicle serving as ground truth annotations. For more detailed information regarding the construction methodology of the fine-tuning dataset, please refer to the Appendix. 

During the fine-tuning stage, we initialize the model with pre-trained weights from the real-time planner. The total training loss comprises two components: (1) the planner's inherent planning loss $\mathcal{L}_{\text{plan}}$; and (2) following AsyncDriver \cite{chen2024asynchronous}, we employ auxiliary tasks including ego vehicle velocity/acceleration prediction, velocity decision prediction, traffic light state prediction, adjacent lane existence prediction, and lane change prediction. The combined loss from all auxiliary tasks is denoted as $\mathcal{L}_{\text{aux}}$. Our composite loss function is formulated as: 
\begin{equation}
    \mathcal{L} = \mathcal{L}_{\text{plan}} + \mathcal{L}_{\text{aux}}
\end{equation}
We implement two distinct versions based on different real-time planners, GameFormer \cite{huang2023gameformer} and PlanTF \cite{cheng2024rethinking}, consequently resulting in slightly divergent planning loss formulations. For the GameFormer implementation, the planning loss consists of two components: (1) $m$ different modes of trajectories of neighbor agents are represented by Gaussian mixture model. For each mode, at any given timestamp $t$, its characteristics are delineated by a mean $\mu^{t}$ and convariance $\sigma^{t}$ forming a Gaussian distribution. The best mode $m^*$ is identified through alignment with the ground truth and refined via the minimization of negative log-likelihood. (2) The ego vehicle's trajectory are predicted and refined by L1 loss. Thus, the complete planning loss formulation is as follows: 
\begin{equation}
    \mathcal{L}_{\text{plan}} = \sum_{t} \Big( \mathcal{L}_{\text{NLL}}(\tilde{\mu}_{m^*}^{t}, \tilde{\sigma}_{m^*}^{t}, \tilde{p}_{m^*}^{t}, 
    \tilde{s}_t) + \mathcal{L}_{1}(\tilde{s}_t, s_t) \Big)
\end{equation}

For the PlanTF implementation, the planning loss comprises three components: (1) The model predicts $m$ trajectory points $s$ with corresponding probabilities $p$ for the ego vehicle at each timestamp $t$, then identifies the optimal trajectory point and its associated mode $m^*$ by comparing with the ground truth trajectory, followed by computing the L1 loss for the trajectory; (2) The cross-entropy loss between the predicted probabilities and the best mode $m^*$; (3) The model additionally predicts trajectories $a$ for all surrounding objects and computes their L1 losses. Thus, the complete planning loss formulation is as follows:
\begin{equation}
    \mathcal{L}_{\text{plan}} = \sum_{t} \Big( \mathcal{L}_{1}(s_{m^*}^{t}, \tilde{s}_t) + \mathcal{L}_{\text{ce}}(p_t, m^*) \Big) + \mathcal{L}_{1}(\tilde{a}, a)
\end{equation}

%% file: Section/4-exp.tex
\section{Experiments}

\subsection{Experimental Setup}
\noindent$\bullet$~\textbf{Evaluation Settings.} Following the nuPlan $2023$ challenge settings \cite{nuplan}, we select $14$ officially designated challenging scenarios for training and evaluation. Since the nuPlan test set contains numerous scenarios where most conventional cases fail to effectively assess critical planning algorithm capabilities, and the massive test volume would significantly prolong evaluation time, 
we construct a long-tail scenario test set following AsyncDriver \cite{chen2024asynchronous}  and PlanTF \cite{cheng2024rethinking}.
First, we randomly sample $100$ scenarios per category across all $14$ scenario types from the test set and evaluate them using PDM (the nuPlan $2023$ champion). Subsequently, we select the $20$ worst-performing scenarios from each category. This selection process is applied consistently to both open-loop and closed-loop testing configurations, resulting in two distinct benchmark datasets: Open-Hard20 and Close-Hard20. 

Furthermore, as our method requires image inputs while nuPlan's closed-loop testing only provides initial frame image information, we adopt an approach similar to NavSim \cite{im2024navsim} for closed-loop evaluation. Specifically, we compress the testing duration to $8$ seconds and perform trajectory prediction solely at the first frame, persistently executing this initial prediction throughout the subsequent period. Extensive experiments in NavSim have demonstrated that this experimental configuration maintains high consistency with standard closed-loop testing protocols. For open-loop evaluation, our testing methodology remains fully consistent with nuPlan's standard open-loop testing procedures.

\noindent$\bullet$ \textbf{Implementation Details.} All experiments are conducted under three distinct settings: closed-loop reactive, closed-loop non-reactive, and open-loop configurations. The simulation operates at a frequency of $10$Hz, with each iteration predicting the ego vehicle's trajectory for an 8-second horizon. For performance evaluation, we employ the official nuPlan metrics for both open-loop and closed-loop assessments. For model settings, our model is built based on LLaVA-v1.5-7b, LoRA was configured with R=8 and alpha=32. We use the AdamW optimizer and warm-up with a decay scheduler with a learning rate of 2e-5. Our experimental evaluation is conducted using two distinct real-time planners: GameFormer \cite{huang2023gameformer} and PlanTF \cite{cheng2024rethinking}. Detailed implementation specifications for both versions are provided in the Appendix.

\noindent $\bullet$ \textbf{Baseline Methods.} Our baseline methods can be categorized into three distinct classes. Rule-based approaches: PDM-Closed \cite{dauner2023parting}; learning-based approaches: PDM-Hybrid \cite{dauner2023parting},GameFormer \cite{huang2023gameformer}, PlanTF \cite{cheng2024rethinking}; PLUTO \cite{cheng2024pluto}, DTPP \cite{huang2024dtpp}; and LLM-based approaches: AsyncDriver \cite{chen2024asynchronous}, LLM-Assist \cite{sharan2023llm}. More details about baseline methods can be found in Appendix.

\subsection{Main Results}

\noindent $\bullet$ \textbf{Open-Hard20 Results.} 
As illustrated in Tab. \ref{tab:open_result}, both versions of our VLMPlanner achieve performance improvements over their respective baseline planners, significantly enhancing the real-time planner's adaptability in complex scenarios. Furthermore, our method outperforms PlanTF \cite{cheng2024rethinking}, the current state-of-the-art open-loop approach. Notably, when implemented on GameFormer \cite{huang2023gameformer}, AsyncDriver \cite{chen2024asynchronous} exhibits degraded performance compared to the original GameFormer in open-loop evaluations.

\noindent$\bullet$ \textbf{Close-Hard20 Results.} As illustrated in Tab. \ref{tab:close_nr_result} and Tab. \ref{tab:close_r_result}, our VLMPlanner achieves state-of-the-art performance across both reactive and non-reactive experimental configurations. Specifically, in the non-reactive setting, it demonstrates a $1.68\%$ performance improvement (approximately $1.26$ score) over DTPP, while under reactive conditions, it achieves a $3.33\%$ enhancement (approximately $2.15$ score) compared to GameFormer. These quantitative results substantiate the effectiveness of our approach in handling complex and challenging scenarios within the nuPlan benchmark.

More detailed metrics reveal that VLMPlanner significantly reduces the collision probability for the ego vehicle. This safety enhancement originates from the real-time planner's improved scene comprehension capability through VLM integration, with particular attention to proximal regions surrounding the ego vehicle, thereby substantially elevating autonomous driving safety.

\noindent $\bullet$ \textbf{Qualitative results.} As depicted in Fig. \ref{fig:visualization}, we provide a comprehensive visual comparison of VLMPlanner and AsyncDriver's behavioral performance across multiple driving scenarios under non-reactive experimental conditions. (a) A "waiting for pedestrian to cross" scenario: As evidenced by the front-left perspective image, a pedestrian has initiated crossing the crosswalk. However, AsyncDriver failed to detect the pedestrian accurately and proceeded to traverse, resulting in a collision with the pedestrian. In contrast, VLMPlanner correctly maintained a stationary position until the pedestrian completed the crossing before proceeding. (b) A "stationary in traffic" scenario:  As captured in the frontal-view image, all directional traffic signals in the current lane are displaying red. However, AsyncDriver failed to properly recognize the traffic signals and proceeded forward, resulting in a collision with the preceding vehicle. In contrast, VLMPlanner correctly chose to wait for the traffic light. These results collectively demonstrate that multi-view image-based scene understanding enables more comprehensive perception of scenario details while effectively preventing potential accidents.

\subsection{Ablation Study}

\noindent $\bullet$ \textbf{Context-Adaptive Inference Gate.} Our CAI-Gate module categorizes scene complexity into five distinct levels and dynamically adjusts the VLM inference intervals accordingly. As illustrated in Tab. \ref{tab:cai-gate}, we systematically evaluated three experimental configurations. setting1: [10, 20, 40, 60, 90]; setting2: [20, 40, 60, 90, 120]; setting3: [40, 60, 90, 120, 160], where each array element corresponds to the inference interval for a specific complexity level. "Average Interval" denotes the mean inference interval measured in Open-Hard20 tests, representing the number of real-time planner inferences between consecutive VLM inferences. "fixed interval" implements periodic VLM inference at fixed intervals (20/60 steps), following AsyncDriver \cite{chen2024asynchronous}. Experimental results demonstrate that  CAI-Gate maintains comparable performance even with reduced VLM inference frequency, , outperforming the fixed-interval inference approach employed in AsyncDriver. And the learning-based CAI-Gate can sustain satisfactory performance even when the inference interval reaches $91$.

\begin{table}[t]
    \caption{Ablation study of CAI-Gate. }
    \label{tab:cai-gate}
        \begin{tabular}{cccc}
            \toprule
            Method & Setting & Score & Average Interval    \\
            \midrule
            GameFormer \cite{huang2023gameformer} & - & $76.22$ & - \\
            \midrule
            \multirow{2}{*}{fixed interval}
            & - & $76.17$ & $20.00$ \\
            & - & $76.16$ & $60.00$ \\
            \midrule
            \multirow{3}{*}{learning-based}
            & setting1 & $76.50$ & $28.79$ \\
            & setting2 & $76.16$ & $40.81$ \\
            & setting3 & $76.16$ & $91.00$ \\
            \midrule
            \multirow{3}{*}{rule-based}
            & setting1 & $76.47$ & $13.76$ \\
            & setting2 & $76.13$ & $50.58$ \\
            & setting3 & $76.00$ & $70.04$ \\
            \midrule
            Ours & - & $\mathbf{76.54}$ & $1.00$ \\
            \bottomrule
        \end{tabular}
\end{table}

\noindent $\bullet$ \textbf{Pretrain.} We further investigate the impact of pretraining on experimental outcomes. As evidenced in Tab. \ref{tab:pretrain_result}, after pretraining with both DriveVQA and ReasoningVQA datasets, significant performance improvements are observed in both reactive and non-reactive closed-loop testing scenarios. This enhancement can be attributed to the effectiveness of the pre-training datasets in elevating the VLM's comprehension and reasoning capabilities regarding lane structures, vehicle/pedestrian interactions, and traffic regulations in autonomous driving contexts.
\begin{table}[t]
    \caption{Ablation study on pretraining effectiveness, where the metrics presented in the table correspond to the average scores.}
    \label{tab:pretrain_result}
        \begin{tabular}{cccc}
            \toprule
            DriveVQA & ReasoningVQA & CLS\_NR & CLS\_R    \\
            \midrule
                        &               & $74.42$ & $65.53$ \\
            \Checkmark  &               & $74.31$ & $66.52$ \\
            \Checkmark  & \Checkmark    & $\mathbf{76.35}$ & $\mathbf{66.66}$ \\
            \bottomrule
        \end{tabular}
\end{table}


%% file: Section/5-conclusion.tex
\section{Conclusions}
In this work, we propose VLMPlanner, an autonomous driving motion planning framework that integrates a learning-based real-time planner with vision-language model (VLM). As illustrated in Fig. \ref{fig:overview}, by incorporating multi-view images as supplementary inputs to the VLM, the system significantly enhances scene comprehension capabilities to handle various long-tail scenarios. Meanwhile, the CAI-Gate module dynamically adjusts the VLM's inference frequency based on scene complexity, achieving an optimal balance between performance and computational efficiency. Experimental results demonstrate that VLMPlanner effectively improves the real-time planner's performance in both open-loop and closed-loop tests on the nuPlan benchmark.

%% file: sample-sigconf.bbl

\begin{thebibliography}{69}


\ifx \showCODEN    \undefined \def \showCODEN     #1{\unskip}     \fi
\ifx \showISBNx    \undefined \def \showISBNx     #1{\unskip}     \fi
\ifx \showISBNxiii \undefined \def \showISBNxiii  #1{\unskip}     \fi
\ifx \showISSN     \undefined \def \showISSN      #1{\unskip}     \fi
\ifx \showLCCN     \undefined \def \showLCCN      #1{\unskip}     \fi
\ifx \shownote     \undefined \def \shownote      #1{#1}          \fi
\ifx \showarticletitle \undefined \def \showarticletitle #1{#1}   \fi
\ifx \showURL      \undefined \def \showURL       {\relax}        \fi
\providecommand\bibfield[2]{#2}
\providecommand\bibinfo[2]{#2}
\providecommand\natexlab[1]{#1}
\providecommand\showeprint[2][]{arXiv:#2}

\bibitem[nup(2019)]%
        {nuplan}
 \bibinfo{year}{2019}\natexlab{}.
\newblock \bibinfo{title}{Motional: nuplan challange.}
\newblock
\urldef\tempurl%
\url{https://github.com/motional/nuplan-devkit}
\showURL{%
\tempurl}


\bibitem[Achiam et~al\mbox{.}(2023a)]%
        {achiam2023gpt4}
\bibfield{author}{\bibinfo{person}{Josh Achiam}, \bibinfo{person}{Steven Adler}, \bibinfo{person}{Sandhini Agarwal}, \bibinfo{person}{Lama Ahmad}, \bibinfo{person}{Ilge Akkaya}, \bibinfo{person}{Florencia~Leoni Aleman}, \bibinfo{person}{Diogo Almeida}, \bibinfo{person}{Janko Altenschmidt}, \bibinfo{person}{Sam Altman}, \bibinfo{person}{Shyamal Anadkat}, {et~al\mbox{.}}} \bibinfo{year}{2023}\natexlab{a}.
\newblock \showarticletitle{Gpt-4 technical report}.
\newblock \bibinfo{journal}{\emph{arXiv preprint arXiv:2303.08774}} (\bibinfo{year}{2023}).
\newblock


\bibitem[Achiam et~al\mbox{.}(2023b)]%
        {achiam2023gpt}
\bibfield{author}{\bibinfo{person}{Josh Achiam}, \bibinfo{person}{Steven Adler}, \bibinfo{person}{Sandhini Agarwal}, \bibinfo{person}{Lama Ahmad}, \bibinfo{person}{Ilge Akkaya}, \bibinfo{person}{Florencia~Leoni Aleman}, \bibinfo{person}{Diogo Almeida}, \bibinfo{person}{Janko Altenschmidt}, \bibinfo{person}{Sam Altman}, \bibinfo{person}{Shyamal Anadkat}, {et~al\mbox{.}}} \bibinfo{year}{2023}\natexlab{b}.
\newblock \showarticletitle{Gpt-4 technical report}.
\newblock \bibinfo{journal}{\emph{arXiv preprint arXiv:2303.08774}} (\bibinfo{year}{2023}).
\newblock


\bibitem[Caesar et~al\mbox{.}(2021)]%
        {caesar2021nuplan}
\bibfield{author}{\bibinfo{person}{Holger Caesar}, \bibinfo{person}{Juraj Kabzan}, \bibinfo{person}{Kok~Seang Tan}, \bibinfo{person}{Whye~Kit Fong}, \bibinfo{person}{Eric Wolff}, \bibinfo{person}{Alex Lang}, \bibinfo{person}{Luke Fletcher}, \bibinfo{person}{Oscar Beijbom}, {and} \bibinfo{person}{Sammy Omari}.} \bibinfo{year}{2021}\natexlab{}.
\newblock \showarticletitle{nuplan: A closed-loop ml-based planning benchmark for autonomous vehicles}.
\newblock \bibinfo{journal}{\emph{arXiv preprint arXiv:2106.11810}} (\bibinfo{year}{2021}).
\newblock


\bibitem[Chen et~al\mbox{.}(2024b)]%
        {chen2024driving}
\bibfield{author}{\bibinfo{person}{Long Chen}, \bibinfo{person}{Oleg Sinavski}, \bibinfo{person}{Jan H{\"u}nermann}, \bibinfo{person}{Alice Karnsund}, \bibinfo{person}{Andrew~James Willmott}, \bibinfo{person}{Danny Birch}, \bibinfo{person}{Daniel Maund}, {and} \bibinfo{person}{Jamie Shotton}.} \bibinfo{year}{2024}\natexlab{b}.
\newblock \showarticletitle{Driving with llms: Fusing object-level vector modality for explainable autonomous driving}. In \bibinfo{booktitle}{\emph{2024 IEEE International Conference on Robotics and Automation (ICRA)}}. IEEE, \bibinfo{pages}{14093--14100}.
\newblock


\bibitem[Chen et~al\mbox{.}(2023)]%
        {chen2023towards}
\bibfield{author}{\bibinfo{person}{Liang Chen}, \bibinfo{person}{Yichi Zhang}, \bibinfo{person}{Shuhuai Ren}, \bibinfo{person}{Haozhe Zhao}, \bibinfo{person}{Zefan Cai}, \bibinfo{person}{Yuchi Wang}, \bibinfo{person}{Peiyi Wang}, \bibinfo{person}{Tianyu Liu}, {and} \bibinfo{person}{Baobao Chang}.} \bibinfo{year}{2023}\natexlab{}.
\newblock \showarticletitle{Towards end-to-end embodied decision making via multi-modal large language model: Explorations with gpt4-vision and beyond}.
\newblock \bibinfo{journal}{\emph{arXiv preprint arXiv:2310.02071}} (\bibinfo{year}{2023}).
\newblock


\bibitem[Chen et~al\mbox{.}(2024a)]%
        {chen2024asynchronous}
\bibfield{author}{\bibinfo{person}{Yuan Chen}, \bibinfo{person}{Zi-han Ding}, \bibinfo{person}{Ziqin Wang}, \bibinfo{person}{Yan Wang}, \bibinfo{person}{Lijun Zhang}, {and} \bibinfo{person}{Si Liu}.} \bibinfo{year}{2024}\natexlab{a}.
\newblock \showarticletitle{Asynchronous large language model enhanced planner for autonomous driving}. In \bibinfo{booktitle}{\emph{European Conference on Computer Vision}}. Springer, \bibinfo{pages}{22--38}.
\newblock


\bibitem[Cheng et~al\mbox{.}(2024a)]%
        {cheng2024pluto}
\bibfield{author}{\bibinfo{person}{Jie Cheng}, \bibinfo{person}{Yingbing Chen}, {and} \bibinfo{person}{Qifeng Chen}.} \bibinfo{year}{2024}\natexlab{a}.
\newblock \showarticletitle{Pluto: Pushing the limit of imitation learning-based planning for autonomous driving}.
\newblock \bibinfo{journal}{\emph{arXiv preprint arXiv:2404.14327}} (\bibinfo{year}{2024}).
\newblock


\bibitem[Cheng et~al\mbox{.}(2024b)]%
        {cheng2024rethinking}
\bibfield{author}{\bibinfo{person}{Jie Cheng}, \bibinfo{person}{Yingbing Chen}, \bibinfo{person}{Xiaodong Mei}, \bibinfo{person}{Bowen Yang}, \bibinfo{person}{Bo Li}, {and} \bibinfo{person}{Ming Liu}.} \bibinfo{year}{2024}\natexlab{b}.
\newblock \showarticletitle{Rethinking imitation-based planners for autonomous driving}. In \bibinfo{booktitle}{\emph{2024 IEEE International Conference on Robotics and Automation (ICRA)}}. IEEE, \bibinfo{pages}{14123--14130}.
\newblock


\bibitem[Chitta et~al\mbox{.}(2022)]%
        {chitta2022transfuser}
\bibfield{author}{\bibinfo{person}{Kashyap Chitta}, \bibinfo{person}{Aditya Prakash}, \bibinfo{person}{Bernhard Jaeger}, \bibinfo{person}{Zehao Yu}, \bibinfo{person}{Katrin Renz}, {and} \bibinfo{person}{Andreas Geiger}.} \bibinfo{year}{2022}\natexlab{}.
\newblock \showarticletitle{Transfuser: Imitation with transformer-based sensor fusion for autonomous driving}.
\newblock \bibinfo{journal}{\emph{IEEE transactions on pattern analysis and machine intelligence}} \bibinfo{volume}{45}, \bibinfo{number}{11} (\bibinfo{year}{2022}), \bibinfo{pages}{12878--12895}.
\newblock


\bibitem[Codevilla et~al\mbox{.}(2019)]%
        {codevilla2019exploring}
\bibfield{author}{\bibinfo{person}{Felipe Codevilla}, \bibinfo{person}{Eder Santana}, \bibinfo{person}{Antonio~M L{\'o}pez}, {and} \bibinfo{person}{Adrien Gaidon}.} \bibinfo{year}{2019}\natexlab{}.
\newblock \showarticletitle{Exploring the limitations of behavior cloning for autonomous driving}. In \bibinfo{booktitle}{\emph{Proceedings of the IEEE/CVF international conference on computer vision}}. \bibinfo{pages}{9329--9338}.
\newblock


\bibitem[Cui et~al\mbox{.}(2024a)]%
        {cui2024receive}
\bibfield{author}{\bibinfo{person}{Can Cui}, \bibinfo{person}{Yunsheng Ma}, \bibinfo{person}{Xu Cao}, \bibinfo{person}{Wenqian Ye}, {and} \bibinfo{person}{Ziran Wang}.} \bibinfo{year}{2024}\natexlab{a}.
\newblock \showarticletitle{Receive, reason, and react: Drive as you say, with large language models in autonomous vehicles}.
\newblock \bibinfo{journal}{\emph{IEEE Intelligent Transportation Systems Magazine}} (\bibinfo{year}{2024}).
\newblock


\bibitem[Cui et~al\mbox{.}(2024b)]%
        {cui2024survey}
\bibfield{author}{\bibinfo{person}{Can Cui}, \bibinfo{person}{Yunsheng Ma}, \bibinfo{person}{Xu Cao}, \bibinfo{person}{Wenqian Ye}, \bibinfo{person}{Yang Zhou}, \bibinfo{person}{Kaizhao Liang}, \bibinfo{person}{Jintai Chen}, \bibinfo{person}{Juanwu Lu}, \bibinfo{person}{Zichong Yang}, \bibinfo{person}{Kuei-Da Liao}, {et~al\mbox{.}}} \bibinfo{year}{2024}\natexlab{b}.
\newblock \showarticletitle{A survey on multimodal large language models for autonomous driving}. In \bibinfo{booktitle}{\emph{Proceedings of the IEEE/CVF Winter Conference on Applications of Computer Vision}}. \bibinfo{pages}{958--979}.
\newblock


\bibitem[Dauner et~al\mbox{.}(2023)]%
        {dauner2023parting}
\bibfield{author}{\bibinfo{person}{Daniel Dauner}, \bibinfo{person}{Marcel Hallgarten}, \bibinfo{person}{Andreas Geiger}, {and} \bibinfo{person}{Kashyap Chitta}.} \bibinfo{year}{2023}\natexlab{}.
\newblock \showarticletitle{Parting with misconceptions about learning-based vehicle motion planning}. In \bibinfo{booktitle}{\emph{Conference on Robot Learning}}. PMLR, \bibinfo{pages}{1268--1281}.
\newblock


\bibitem[Ding et~al\mbox{.}(2024)]%
        {ding2024holistic}
\bibfield{author}{\bibinfo{person}{Xinpeng Ding}, \bibinfo{person}{Jianhua Han}, \bibinfo{person}{Hang Xu}, \bibinfo{person}{Xiaodan Liang}, \bibinfo{person}{Wei Zhang}, {and} \bibinfo{person}{Xiaomeng Li}.} \bibinfo{year}{2024}\natexlab{}.
\newblock \showarticletitle{Holistic autonomous driving understanding by bird's-eye-view injected multi-modal large models}. In \bibinfo{booktitle}{\emph{Proceedings of the IEEE/CVF Conference on Computer Vision and Pattern Recognition}}. \bibinfo{pages}{13668--13677}.
\newblock


\bibitem[Fu et~al\mbox{.}(2024)]%
        {fu2024drive}
\bibfield{author}{\bibinfo{person}{Daocheng Fu}, \bibinfo{person}{Xin Li}, \bibinfo{person}{Licheng Wen}, \bibinfo{person}{Min Dou}, \bibinfo{person}{Pinlong Cai}, \bibinfo{person}{Botian Shi}, {and} \bibinfo{person}{Yu Qiao}.} \bibinfo{year}{2024}\natexlab{}.
\newblock \showarticletitle{Drive like a human: Rethinking autonomous driving with large language models}. In \bibinfo{booktitle}{\emph{2024 IEEE/CVF Winter Conference on Applications of Computer Vision Workshops (WACVW)}}. IEEE, \bibinfo{pages}{910--919}.
\newblock


\bibitem[Grattafiori et~al\mbox{.}(2024)]%
        {grattafiori2024llama}
\bibfield{author}{\bibinfo{person}{Aaron Grattafiori}, \bibinfo{person}{Abhimanyu Dubey}, \bibinfo{person}{Abhinav Jauhri}, \bibinfo{person}{Abhinav Pandey}, \bibinfo{person}{Abhishek Kadian}, \bibinfo{person}{Ahmad Al-Dahle}, \bibinfo{person}{Aiesha Letman}, \bibinfo{person}{Akhil Mathur}, \bibinfo{person}{Alan Schelten}, \bibinfo{person}{Alex Vaughan}, {et~al\mbox{.}}} \bibinfo{year}{2024}\natexlab{}.
\newblock \showarticletitle{The llama 3 herd of models}.
\newblock \bibinfo{journal}{\emph{arXiv preprint arXiv:2407.21783}} (\bibinfo{year}{2024}).
\newblock


\bibitem[Hallgarten et~al\mbox{.}(2023)]%
        {hallgarten2023prediction}
\bibfield{author}{\bibinfo{person}{Marcel Hallgarten}, \bibinfo{person}{Martin Stoll}, {and} \bibinfo{person}{Andreas Zell}.} \bibinfo{year}{2023}\natexlab{}.
\newblock \showarticletitle{From prediction to planning with goal conditioned lane graph traversals}. In \bibinfo{booktitle}{\emph{2023 IEEE 26th International Conference on Intelligent Transportation Systems (ITSC)}}. IEEE, \bibinfo{pages}{951--958}.
\newblock


\bibitem[Han et~al\mbox{.}(2024)]%
        {han2024dme}
\bibfield{author}{\bibinfo{person}{Wencheng Han}, \bibinfo{person}{Dongqian Guo}, \bibinfo{person}{Cheng-Zhong Xu}, {and} \bibinfo{person}{Jianbing Shen}.} \bibinfo{year}{2024}\natexlab{}.
\newblock \showarticletitle{Dme-driver: Integrating human decision logic and 3d scene perception in autonomous driving}.
\newblock \bibinfo{journal}{\emph{arXiv preprint arXiv:2401.03641}} (\bibinfo{year}{2024}).
\newblock


\bibitem[Hu et~al\mbox{.}(2022)]%
        {hu2022st}
\bibfield{author}{\bibinfo{person}{Shengchao Hu}, \bibinfo{person}{Li Chen}, \bibinfo{person}{Penghao Wu}, \bibinfo{person}{Hongyang Li}, \bibinfo{person}{Junchi Yan}, {and} \bibinfo{person}{Dacheng Tao}.} \bibinfo{year}{2022}\natexlab{}.
\newblock \showarticletitle{St-p3: End-to-end vision-based autonomous driving via spatial-temporal feature learning}. In \bibinfo{booktitle}{\emph{European Conference on Computer Vision}}. Springer, \bibinfo{pages}{533--549}.
\newblock


\bibitem[Hu et~al\mbox{.}(2024)]%
        {hu2024solving}
\bibfield{author}{\bibinfo{person}{Yihan Hu}, \bibinfo{person}{Siqi Chai}, \bibinfo{person}{Zhening Yang}, \bibinfo{person}{Jingyu Qian}, \bibinfo{person}{Kun Li}, \bibinfo{person}{Wenxin Shao}, \bibinfo{person}{Haichao Zhang}, \bibinfo{person}{Wei Xu}, {and} \bibinfo{person}{Qiang Liu}.} \bibinfo{year}{2024}\natexlab{}.
\newblock \showarticletitle{Solving motion planning tasks with a scalable generative model}. In \bibinfo{booktitle}{\emph{European Conference on Computer Vision}}. Springer, \bibinfo{pages}{386--404}.
\newblock


\bibitem[Hu et~al\mbox{.}(2023a)]%
        {hu2023imitation}
\bibfield{author}{\bibinfo{person}{Yihan Hu}, \bibinfo{person}{Kun Li}, \bibinfo{person}{Pingyuan Liang}, \bibinfo{person}{Jingyu Qian}, \bibinfo{person}{Zhening Yang}, \bibinfo{person}{Haichao Zhang}, \bibinfo{person}{Wenxin Shao}, \bibinfo{person}{Zhuangzhuang Ding}, \bibinfo{person}{Wei Xu}, {and} \bibinfo{person}{Qiang Liu}.} \bibinfo{year}{2023}\natexlab{a}.
\newblock \showarticletitle{Imitation with spatial-temporal heatmap: 2nd place solution for nuplan challenge}.
\newblock \bibinfo{journal}{\emph{arXiv preprint arXiv:2306.15700}} (\bibinfo{year}{2023}).
\newblock


\bibitem[Hu et~al\mbox{.}(2023b)]%
        {hu2023planning}
\bibfield{author}{\bibinfo{person}{Yihan Hu}, \bibinfo{person}{Jiazhi Yang}, \bibinfo{person}{Li Chen}, \bibinfo{person}{Keyu Li}, \bibinfo{person}{Chonghao Sima}, \bibinfo{person}{Xizhou Zhu}, \bibinfo{person}{Siqi Chai}, \bibinfo{person}{Senyao Du}, \bibinfo{person}{Tianwei Lin}, \bibinfo{person}{Wenhai Wang}, {et~al\mbox{.}}} \bibinfo{year}{2023}\natexlab{b}.
\newblock \showarticletitle{Planning-oriented autonomous driving}. In \bibinfo{booktitle}{\emph{Proceedings of the IEEE/CVF conference on computer vision and pattern recognition}}. \bibinfo{pages}{17853--17862}.
\newblock


\bibitem[Huang et~al\mbox{.}(2024)]%
        {huang2024dtpp}
\bibfield{author}{\bibinfo{person}{Zhiyu Huang}, \bibinfo{person}{Peter Karkus}, \bibinfo{person}{Boris Ivanovic}, \bibinfo{person}{Yuxiao Chen}, \bibinfo{person}{Marco Pavone}, {and} \bibinfo{person}{Chen Lv}.} \bibinfo{year}{2024}\natexlab{}.
\newblock \showarticletitle{Dtpp: Differentiable joint conditional prediction and cost evaluation for tree policy planning in autonomous driving}. In \bibinfo{booktitle}{\emph{2024 IEEE International Conference on Robotics and Automation (ICRA)}}. IEEE, \bibinfo{pages}{6806--6812}.
\newblock


\bibitem[Huang et~al\mbox{.}(2023a)]%
        {huang2023gameformer}
\bibfield{author}{\bibinfo{person}{Zhiyu Huang}, \bibinfo{person}{Haochen Liu}, {and} \bibinfo{person}{Chen Lv}.} \bibinfo{year}{2023}\natexlab{a}.
\newblock \showarticletitle{Gameformer: Game-theoretic modeling and learning of transformer-based interactive prediction and planning for autonomous driving}. In \bibinfo{booktitle}{\emph{Proceedings of the IEEE/CVF International Conference on Computer Vision}}. \bibinfo{pages}{3903--3913}.
\newblock


\bibitem[Huang et~al\mbox{.}(2023b)]%
        {huang2023differentiable}
\bibfield{author}{\bibinfo{person}{Zhiyu Huang}, \bibinfo{person}{Haochen Liu}, \bibinfo{person}{Jingda Wu}, {and} \bibinfo{person}{Chen Lv}.} \bibinfo{year}{2023}\natexlab{b}.
\newblock \showarticletitle{Differentiable integrated motion prediction and planning with learnable cost function for autonomous driving}.
\newblock \bibinfo{journal}{\emph{IEEE transactions on neural networks and learning systems}} (\bibinfo{year}{2023}).
\newblock


\bibitem[Hwang et~al\mbox{.}(2024)]%
        {hwang2024emma}
\bibfield{author}{\bibinfo{person}{Jyh-Jing Hwang}, \bibinfo{person}{Runsheng Xu}, \bibinfo{person}{Hubert Lin}, \bibinfo{person}{Wei-Chih Hung}, \bibinfo{person}{Jingwei Ji}, \bibinfo{person}{Kristy Choi}, \bibinfo{person}{Di Huang}, \bibinfo{person}{Tong He}, \bibinfo{person}{Paul Covington}, \bibinfo{person}{Benjamin Sapp}, {et~al\mbox{.}}} \bibinfo{year}{2024}\natexlab{}.
\newblock \showarticletitle{Emma: End-to-end multimodal model for autonomous driving}.
\newblock \bibinfo{journal}{\emph{arXiv preprint arXiv:2410.23262}} (\bibinfo{year}{2024}).
\newblock


\bibitem[IM(2024)]%
        {im2024navsim}
\bibfield{author}{\bibinfo{person}{NAV IM}.} \bibinfo{year}{2024}\natexlab{}.
\newblock \showarticletitle{NAVSIM: Data-Driven Non-Reactive Autonomous Vehicle Simulation and Benchmarking}.
\newblock \bibinfo{journal}{\emph{arXiv preprint arXiv:2406.15349}} (\bibinfo{year}{2024}).
\newblock


\bibitem[Jia et~al\mbox{.}(2023)]%
        {jia2023think}
\bibfield{author}{\bibinfo{person}{Xiaosong Jia}, \bibinfo{person}{Penghao Wu}, \bibinfo{person}{Li Chen}, \bibinfo{person}{Jiangwei Xie}, \bibinfo{person}{Conghui He}, \bibinfo{person}{Junchi Yan}, {and} \bibinfo{person}{Hongyang Li}.} \bibinfo{year}{2023}\natexlab{}.
\newblock \showarticletitle{Think twice before driving: Towards scalable decoders for end-to-end autonomous driving}. In \bibinfo{booktitle}{\emph{Proceedings of the IEEE/CVF Conference on Computer Vision and Pattern Recognition}}. \bibinfo{pages}{21983--21994}.
\newblock


\bibitem[Jiang et~al\mbox{.}(2024)]%
        {jiang2024senna}
\bibfield{author}{\bibinfo{person}{Bo Jiang}, \bibinfo{person}{Shaoyu Chen}, \bibinfo{person}{Bencheng Liao}, \bibinfo{person}{Xingyu Zhang}, \bibinfo{person}{Wei Yin}, \bibinfo{person}{Qian Zhang}, \bibinfo{person}{Chang Huang}, \bibinfo{person}{Wenyu Liu}, {and} \bibinfo{person}{Xinggang Wang}.} \bibinfo{year}{2024}\natexlab{}.
\newblock \showarticletitle{Senna: Bridging large vision-language models and end-to-end autonomous driving}.
\newblock \bibinfo{journal}{\emph{arXiv preprint arXiv:2410.22313}} (\bibinfo{year}{2024}).
\newblock


\bibitem[Jiang et~al\mbox{.}(2023)]%
        {jiang2023vad}
\bibfield{author}{\bibinfo{person}{Bo Jiang}, \bibinfo{person}{Shaoyu Chen}, \bibinfo{person}{Qing Xu}, \bibinfo{person}{Bencheng Liao}, \bibinfo{person}{Jiajie Chen}, \bibinfo{person}{Helong Zhou}, \bibinfo{person}{Qian Zhang}, \bibinfo{person}{Wenyu Liu}, \bibinfo{person}{Chang Huang}, {and} \bibinfo{person}{Xinggang Wang}.} \bibinfo{year}{2023}\natexlab{}.
\newblock \showarticletitle{Vad: Vectorized scene representation for efficient autonomous driving}. In \bibinfo{booktitle}{\emph{Proceedings of the IEEE/CVF International Conference on Computer Vision}}. \bibinfo{pages}{8340--8350}.
\newblock


\bibitem[Jin et~al\mbox{.}(2023)]%
        {jin2023surrealdriver}
\bibfield{author}{\bibinfo{person}{Ye Jin}, \bibinfo{person}{Xiaoxi Shen}, \bibinfo{person}{Huiling Peng}, \bibinfo{person}{Xiaoan Liu}, \bibinfo{person}{Jingli Qin}, \bibinfo{person}{Jiayang Li}, \bibinfo{person}{Jintao Xie}, \bibinfo{person}{Peizhong Gao}, \bibinfo{person}{Guyue Zhou}, {and} \bibinfo{person}{Jiangtao Gong}.} \bibinfo{year}{2023}\natexlab{}.
\newblock \showarticletitle{Surrealdriver: Designing generative driver agent simulation framework in urban contexts based on large language model}.
\newblock \bibinfo{journal}{\emph{arXiv preprint arXiv:2309.13193}} \bibinfo{volume}{5}, \bibinfo{number}{7} (\bibinfo{year}{2023}), \bibinfo{pages}{8}.
\newblock


\bibitem[Kendall et~al\mbox{.}(2019)]%
        {kendall2019learning}
\bibfield{author}{\bibinfo{person}{Alex Kendall}, \bibinfo{person}{Jeffrey Hawke}, \bibinfo{person}{David Janz}, \bibinfo{person}{Przemyslaw Mazur}, \bibinfo{person}{Daniele Reda}, \bibinfo{person}{John-Mark Allen}, \bibinfo{person}{Vinh-Dieu Lam}, \bibinfo{person}{Alex Bewley}, {and} \bibinfo{person}{Amar Shah}.} \bibinfo{year}{2019}\natexlab{}.
\newblock \showarticletitle{Learning to drive in a day}. In \bibinfo{booktitle}{\emph{2019 international conference on robotics and automation (ICRA)}}. IEEE, \bibinfo{pages}{8248--8254}.
\newblock


\bibitem[Kircher and Ahlstrom(2017)]%
        {kircher2017minimum}
\bibfield{author}{\bibinfo{person}{Katja Kircher} {and} \bibinfo{person}{Christer Ahlstrom}.} \bibinfo{year}{2017}\natexlab{}.
\newblock \showarticletitle{Minimum required attention: A human-centered approach to driver inattention}.
\newblock \bibinfo{journal}{\emph{Human factors}} \bibinfo{volume}{59}, \bibinfo{number}{3} (\bibinfo{year}{2017}), \bibinfo{pages}{471--484}.
\newblock


\bibitem[Li et~al\mbox{.}(2022)]%
        {li2022blip}
\bibfield{author}{\bibinfo{person}{Junnan Li}, \bibinfo{person}{Dongxu Li}, \bibinfo{person}{Caiming Xiong}, {and} \bibinfo{person}{Steven Hoi}.} \bibinfo{year}{2022}\natexlab{}.
\newblock \showarticletitle{Blip: Bootstrapping language-image pre-training for unified vision-language understanding and generation}. In \bibinfo{booktitle}{\emph{International conference on machine learning}}. PMLR, \bibinfo{pages}{12888--12900}.
\newblock


\bibitem[Li et~al\mbox{.}(2025)]%
        {li2025generative}
\bibfield{author}{\bibinfo{person}{Tengpeng Li}, \bibinfo{person}{Hanli Wang}, \bibinfo{person}{Xianfei Li}, \bibinfo{person}{Wenlong Liao}, \bibinfo{person}{Tao He}, {and} \bibinfo{person}{Pai Peng}.} \bibinfo{year}{2025}\natexlab{}.
\newblock \showarticletitle{Generative Planning with 3D-vision Language Pre-training for End-to-End Autonomous Driving}.
\newblock \bibinfo{journal}{\emph{arXiv preprint arXiv:2501.08861}} (\bibinfo{year}{2025}).
\newblock


\bibitem[Li et~al\mbox{.}(2024)]%
        {li2024boosting}
\bibfield{author}{\bibinfo{person}{Zenan Li}, \bibinfo{person}{Fan Nie}, \bibinfo{person}{Qiao Sun}, \bibinfo{person}{Fang Da}, {and} \bibinfo{person}{Hang Zhao}.} \bibinfo{year}{2024}\natexlab{}.
\newblock \showarticletitle{Boosting offline reinforcement learning for autonomous driving with hierarchical latent skills}. In \bibinfo{booktitle}{\emph{2024 IEEE International Conference on Robotics and Automation (ICRA)}}. IEEE, \bibinfo{pages}{18362--18369}.
\newblock


\bibitem[Liu et~al\mbox{.}(2023)]%
        {liu2023mtd}
\bibfield{author}{\bibinfo{person}{Jiaqi Liu}, \bibinfo{person}{Peng Hang}, \bibinfo{person}{Xiao Qi}, \bibinfo{person}{Jianqiang Wang}, {and} \bibinfo{person}{Jian Sun}.} \bibinfo{year}{2023}\natexlab{}.
\newblock \showarticletitle{Mtd-gpt: A multi-task decision-making gpt model for autonomous driving at unsignalized intersections}. In \bibinfo{booktitle}{\emph{2023 IEEE 26th International Conference on Intelligent Transportation Systems (ITSC)}}. IEEE, \bibinfo{pages}{5154--5161}.
\newblock


\bibitem[Liu et~al\mbox{.}(2022)]%
        {liu2022petr}
\bibfield{author}{\bibinfo{person}{Yingfei Liu}, \bibinfo{person}{Tiancai Wang}, \bibinfo{person}{Xiangyu Zhang}, {and} \bibinfo{person}{Jian Sun}.} \bibinfo{year}{2022}\natexlab{}.
\newblock \showarticletitle{Petr: Position embedding transformation for multi-view 3d object detection}. In \bibinfo{booktitle}{\emph{European conference on computer vision}}. Springer, \bibinfo{pages}{531--548}.
\newblock


\bibitem[Liu et~al\mbox{.}(2021)]%
        {liu2021drivers}
\bibfield{author}{\bibinfo{person}{Zhuofan Liu}, \bibinfo{person}{Wei Yuan}, {and} \bibinfo{person}{Yong Ma}.} \bibinfo{year}{2021}\natexlab{}.
\newblock \showarticletitle{Drivers’ attention strategies before eyes-off-road in different traffic scenarios: adaptation and anticipation}.
\newblock \bibinfo{journal}{\emph{International journal of environmental research and public health}} \bibinfo{volume}{18}, \bibinfo{number}{7} (\bibinfo{year}{2021}), \bibinfo{pages}{3716}.
\newblock


\bibitem[Ma et~al\mbox{.}(2024a)]%
        {ma2024dolphins}
\bibfield{author}{\bibinfo{person}{Yingzi Ma}, \bibinfo{person}{Yulong Cao}, \bibinfo{person}{Jiachen Sun}, \bibinfo{person}{Marco Pavone}, {and} \bibinfo{person}{Chaowei Xiao}.} \bibinfo{year}{2024}\natexlab{a}.
\newblock \showarticletitle{Dolphins: Multimodal language model for driving}. In \bibinfo{booktitle}{\emph{European Conference on Computer Vision}}. Springer, \bibinfo{pages}{403--420}.
\newblock


\bibitem[Ma et~al\mbox{.}(2024b)]%
        {ma2024lampilot}
\bibfield{author}{\bibinfo{person}{Yunsheng Ma}, \bibinfo{person}{Can Cui}, \bibinfo{person}{Xu Cao}, \bibinfo{person}{Wenqian Ye}, \bibinfo{person}{Peiran Liu}, \bibinfo{person}{Juanwu Lu}, \bibinfo{person}{Amr Abdelraouf}, \bibinfo{person}{Rohit Gupta}, \bibinfo{person}{Kyungtae Han}, \bibinfo{person}{Aniket Bera}, {et~al\mbox{.}}} \bibinfo{year}{2024}\natexlab{b}.
\newblock \showarticletitle{Lampilot: An open benchmark dataset for autonomous driving with language model programs}. In \bibinfo{booktitle}{\emph{Proceedings of the IEEE/CVF Conference on Computer Vision and Pattern Recognition}}. \bibinfo{pages}{15141--15151}.
\newblock


\bibitem[Mao et~al\mbox{.}(2023a)]%
        {mao2023gpt}
\bibfield{author}{\bibinfo{person}{Jiageng Mao}, \bibinfo{person}{Yuxi Qian}, \bibinfo{person}{Junjie Ye}, \bibinfo{person}{Hang Zhao}, {and} \bibinfo{person}{Yue Wang}.} \bibinfo{year}{2023}\natexlab{a}.
\newblock \showarticletitle{Gpt-driver: Learning to drive with gpt}.
\newblock \bibinfo{journal}{\emph{arXiv preprint arXiv:2310.01415}} (\bibinfo{year}{2023}).
\newblock


\bibitem[Mao et~al\mbox{.}(2023b)]%
        {mao2023language}
\bibfield{author}{\bibinfo{person}{Jiageng Mao}, \bibinfo{person}{Junjie Ye}, \bibinfo{person}{Yuxi Qian}, \bibinfo{person}{Marco Pavone}, {and} \bibinfo{person}{Yue Wang}.} \bibinfo{year}{2023}\natexlab{b}.
\newblock \showarticletitle{A language agent for autonomous driving}.
\newblock \bibinfo{journal}{\emph{arXiv preprint arXiv:2311.10813}} (\bibinfo{year}{2023}).
\newblock


\bibitem[Nie et~al\mbox{.}(2024)]%
        {nie2024reason2drive}
\bibfield{author}{\bibinfo{person}{Ming Nie}, \bibinfo{person}{Renyuan Peng}, \bibinfo{person}{Chunwei Wang}, \bibinfo{person}{Xinyue Cai}, \bibinfo{person}{Jianhua Han}, \bibinfo{person}{Hang Xu}, {and} \bibinfo{person}{Li Zhang}.} \bibinfo{year}{2024}\natexlab{}.
\newblock \showarticletitle{Reason2drive: Towards interpretable and chain-based reasoning for autonomous driving}. In \bibinfo{booktitle}{\emph{European Conference on Computer Vision}}. Springer, \bibinfo{pages}{292--308}.
\newblock


\bibitem[Radford et~al\mbox{.}(2021)]%
        {radford2021learning}
\bibfield{author}{\bibinfo{person}{Alec Radford}, \bibinfo{person}{Jong~Wook Kim}, \bibinfo{person}{Chris Hallacy}, \bibinfo{person}{Aditya Ramesh}, \bibinfo{person}{Gabriel Goh}, \bibinfo{person}{Sandhini Agarwal}, \bibinfo{person}{Girish Sastry}, \bibinfo{person}{Amanda Askell}, \bibinfo{person}{Pamela Mishkin}, \bibinfo{person}{Jack Clark}, {et~al\mbox{.}}} \bibinfo{year}{2021}\natexlab{}.
\newblock \showarticletitle{Learning transferable visual models from natural language supervision}. In \bibinfo{booktitle}{\emph{International conference on machine learning}}. PmLR, \bibinfo{pages}{8748--8763}.
\newblock


\bibitem[Renz et~al\mbox{.}(2022)]%
        {renz2022plant}
\bibfield{author}{\bibinfo{person}{Katrin Renz}, \bibinfo{person}{Kashyap Chitta}, \bibinfo{person}{Otniel-Bogdan Mercea}, \bibinfo{person}{A Koepke}, \bibinfo{person}{Zeynep Akata}, {and} \bibinfo{person}{Andreas Geiger}.} \bibinfo{year}{2022}\natexlab{}.
\newblock \showarticletitle{Plant: Explainable planning transformers via object-level representations}.
\newblock \bibinfo{journal}{\emph{arXiv preprint arXiv:2210.14222}} (\bibinfo{year}{2022}).
\newblock


\bibitem[Scheel et~al\mbox{.}(2022)]%
        {scheel2022urban}
\bibfield{author}{\bibinfo{person}{Oliver Scheel}, \bibinfo{person}{Luca Bergamini}, \bibinfo{person}{Maciej Wolczyk}, \bibinfo{person}{B{\l}a{\.z}ej Osi{\'n}ski}, {and} \bibinfo{person}{Peter Ondruska}.} \bibinfo{year}{2022}\natexlab{}.
\newblock \showarticletitle{Urban driver: Learning to drive from real-world demonstrations using policy gradients}. In \bibinfo{booktitle}{\emph{Conference on Robot Learning}}. PMLR, \bibinfo{pages}{718--728}.
\newblock


\bibitem[Sha et~al\mbox{.}(2023)]%
        {sha2023languagempc}
\bibfield{author}{\bibinfo{person}{Hao Sha}, \bibinfo{person}{Yao Mu}, \bibinfo{person}{Yuxuan Jiang}, \bibinfo{person}{Li Chen}, \bibinfo{person}{Chenfeng Xu}, \bibinfo{person}{Ping Luo}, \bibinfo{person}{Shengbo~Eben Li}, \bibinfo{person}{Masayoshi Tomizuka}, \bibinfo{person}{Wei Zhan}, {and} \bibinfo{person}{Mingyu Ding}.} \bibinfo{year}{2023}\natexlab{}.
\newblock \showarticletitle{Languagempc: Large language models as decision makers for autonomous driving}.
\newblock \bibinfo{journal}{\emph{arXiv preprint arXiv:2310.03026}} (\bibinfo{year}{2023}).
\newblock


\bibitem[Shao et~al\mbox{.}(2024)]%
        {shao2024lmdrive}
\bibfield{author}{\bibinfo{person}{Hao Shao}, \bibinfo{person}{Yuxuan Hu}, \bibinfo{person}{Letian Wang}, \bibinfo{person}{Guanglu Song}, \bibinfo{person}{Steven~L Waslander}, \bibinfo{person}{Yu Liu}, {and} \bibinfo{person}{Hongsheng Li}.} \bibinfo{year}{2024}\natexlab{}.
\newblock \showarticletitle{Lmdrive: Closed-loop end-to-end driving with large language models}. In \bibinfo{booktitle}{\emph{Proceedings of the IEEE/CVF Conference on Computer Vision and Pattern Recognition}}. \bibinfo{pages}{15120--15130}.
\newblock


\bibitem[Sharan et~al\mbox{.}(2023)]%
        {sharan2023llm}
\bibfield{author}{\bibinfo{person}{SP Sharan}, \bibinfo{person}{Francesco Pittaluga}, \bibinfo{person}{Manmohan Chandraker}, {et~al\mbox{.}}} \bibinfo{year}{2023}\natexlab{}.
\newblock \showarticletitle{Llm-assist: Enhancing closed-loop planning with language-based reasoning}.
\newblock \bibinfo{journal}{\emph{arXiv preprint arXiv:2401.00125}} (\bibinfo{year}{2023}).
\newblock


\bibitem[Sima et~al\mbox{.}(2024)]%
        {sima2024drivelm}
\bibfield{author}{\bibinfo{person}{Chonghao Sima}, \bibinfo{person}{Katrin Renz}, \bibinfo{person}{Kashyap Chitta}, \bibinfo{person}{Li Chen}, \bibinfo{person}{Hanxue Zhang}, \bibinfo{person}{Chengen Xie}, \bibinfo{person}{Jens Bei{\ss}wenger}, \bibinfo{person}{Ping Luo}, \bibinfo{person}{Andreas Geiger}, {and} \bibinfo{person}{Hongyang Li}.} \bibinfo{year}{2024}\natexlab{}.
\newblock \showarticletitle{Drivelm: Driving with graph visual question answering}. In \bibinfo{booktitle}{\emph{European Conference on Computer Vision}}. Springer, \bibinfo{pages}{256--274}.
\newblock


\bibitem[Tan and Le(2019)]%
        {tan2019efficientnet}
\bibfield{author}{\bibinfo{person}{Mingxing Tan} {and} \bibinfo{person}{Quoc Le}.} \bibinfo{year}{2019}\natexlab{}.
\newblock \showarticletitle{Efficientnet: Rethinking model scaling for convolutional neural networks}. In \bibinfo{booktitle}{\emph{International conference on machine learning}}. PMLR, \bibinfo{pages}{6105--6114}.
\newblock


\bibitem[Wang et~al\mbox{.}(2024a)]%
        {wang2024lhpf}
\bibfield{author}{\bibinfo{person}{Sheng Wang}, \bibinfo{person}{Yao Tian}, \bibinfo{person}{Xiaodong Mei}, \bibinfo{person}{Ge Sun}, \bibinfo{person}{Jie Cheng}, \bibinfo{person}{Fulong Ma}, \bibinfo{person}{Pedro~V Sander}, {and} \bibinfo{person}{Junwei Liang}.} \bibinfo{year}{2024}\natexlab{a}.
\newblock \showarticletitle{LHPF: Look back the History and Plan for the Future in Autonomous Driving}.
\newblock \bibinfo{journal}{\emph{arXiv preprint arXiv:2411.17253}} (\bibinfo{year}{2024}).
\newblock


\bibitem[Wang et~al\mbox{.}(2024b)]%
        {wang2024omnidrive}
\bibfield{author}{\bibinfo{person}{Shihao Wang}, \bibinfo{person}{Zhiding Yu}, \bibinfo{person}{Xiaohui Jiang}, \bibinfo{person}{Shiyi Lan}, \bibinfo{person}{Min Shi}, \bibinfo{person}{Nadine Chang}, \bibinfo{person}{Jan Kautz}, \bibinfo{person}{Ying Li}, {and} \bibinfo{person}{Jose~M Alvarez}.} \bibinfo{year}{2024}\natexlab{b}.
\newblock \showarticletitle{Omnidrive: A holistic llm-agent framework for autonomous driving with 3d perception, reasoning and planning}.
\newblock \bibinfo{journal}{\emph{arXiv preprint arXiv:2405.01533}} (\bibinfo{year}{2024}).
\newblock


\bibitem[Wang et~al\mbox{.}(2023c)]%
        {wang2023chatgpt}
\bibfield{author}{\bibinfo{person}{Shiyi Wang}, \bibinfo{person}{Yuxuan Zhu}, \bibinfo{person}{Zhiheng Li}, \bibinfo{person}{Yutong Wang}, \bibinfo{person}{Li Li}, {and} \bibinfo{person}{Zhengbing He}.} \bibinfo{year}{2023}\natexlab{c}.
\newblock \showarticletitle{Chatgpt as your vehicle co-pilot: An initial attempt}.
\newblock \bibinfo{journal}{\emph{IEEE Transactions on Intelligent Vehicles}} \bibinfo{volume}{8}, \bibinfo{number}{12} (\bibinfo{year}{2023}), \bibinfo{pages}{4706--4721}.
\newblock


\bibitem[Wang et~al\mbox{.}(2023b)]%
        {wang2023drivemlm}
\bibfield{author}{\bibinfo{person}{Wenhai Wang}, \bibinfo{person}{Jiangwei Xie}, \bibinfo{person}{ChuanYang Hu}, \bibinfo{person}{Haoming Zou}, \bibinfo{person}{Jianan Fan}, \bibinfo{person}{Wenwen Tong}, \bibinfo{person}{Yang Wen}, \bibinfo{person}{Silei Wu}, \bibinfo{person}{Hanming Deng}, \bibinfo{person}{Zhiqi Li}, {et~al\mbox{.}}} \bibinfo{year}{2023}\natexlab{b}.
\newblock \showarticletitle{Drivemlm: Aligning multi-modal large language models with behavioral planning states for autonomous driving}.
\newblock \bibinfo{journal}{\emph{arXiv preprint arXiv:2312.09245}} (\bibinfo{year}{2023}).
\newblock


\bibitem[Wang et~al\mbox{.}(2023a)]%
        {wang2023empowering}
\bibfield{author}{\bibinfo{person}{Yixuan Wang}, \bibinfo{person}{Ruochen Jiao}, \bibinfo{person}{Sinong~Simon Zhan}, \bibinfo{person}{Chengtian Lang}, \bibinfo{person}{Chao Huang}, \bibinfo{person}{Zhaoran Wang}, \bibinfo{person}{Zhuoran Yang}, {and} \bibinfo{person}{Qi Zhu}.} \bibinfo{year}{2023}\natexlab{a}.
\newblock \showarticletitle{Empowering autonomous driving with large language models: A safety perspective}.
\newblock \bibinfo{journal}{\emph{arXiv preprint arXiv:2312.00812}} (\bibinfo{year}{2023}).
\newblock


\bibitem[Wen et~al\mbox{.}(2023)]%
        {wen2023dilu}
\bibfield{author}{\bibinfo{person}{Licheng Wen}, \bibinfo{person}{Daocheng Fu}, \bibinfo{person}{Xin Li}, \bibinfo{person}{Xinyu Cai}, \bibinfo{person}{Tao Ma}, \bibinfo{person}{Pinlong Cai}, \bibinfo{person}{Min Dou}, \bibinfo{person}{Botian Shi}, \bibinfo{person}{Liang He}, {and} \bibinfo{person}{Yu Qiao}.} \bibinfo{year}{2023}\natexlab{}.
\newblock \showarticletitle{Dilu: A knowledge-driven approach to autonomous driving with large language models}.
\newblock \bibinfo{journal}{\emph{arXiv preprint arXiv:2309.16292}} (\bibinfo{year}{2023}).
\newblock


\bibitem[Xu et~al\mbox{.}(2024a)]%
        {xu2024vlm}
\bibfield{author}{\bibinfo{person}{Yi Xu}, \bibinfo{person}{Yuxin Hu}, \bibinfo{person}{Zaiwei Zhang}, \bibinfo{person}{Gregory~P Meyer}, \bibinfo{person}{Siva~Karthik Mustikovela}, \bibinfo{person}{Siddhartha Srinivasa}, \bibinfo{person}{Eric~M Wolff}, {and} \bibinfo{person}{Xin Huang}.} \bibinfo{year}{2024}\natexlab{a}.
\newblock \showarticletitle{Vlm-ad: End-to-end autonomous driving through vision-language model supervision}.
\newblock \bibinfo{journal}{\emph{arXiv preprint arXiv:2412.14446}} (\bibinfo{year}{2024}).
\newblock


\bibitem[Xu et~al\mbox{.}(2024b)]%
        {xu2024drivegpt4}
\bibfield{author}{\bibinfo{person}{Zhenhua Xu}, \bibinfo{person}{Yujia Zhang}, \bibinfo{person}{Enze Xie}, \bibinfo{person}{Zhen Zhao}, \bibinfo{person}{Yong Guo}, \bibinfo{person}{Kwan-Yee~K Wong}, \bibinfo{person}{Zhenguo Li}, {and} \bibinfo{person}{Hengshuang Zhao}.} \bibinfo{year}{2024}\natexlab{b}.
\newblock \showarticletitle{Drivegpt4: Interpretable end-to-end autonomous driving via large language model}.
\newblock \bibinfo{journal}{\emph{IEEE Robotics and Automation Letters}} (\bibinfo{year}{2024}).
\newblock


\bibitem[Yang et~al\mbox{.}(2024)]%
        {yang2024diffusion}
\bibfield{author}{\bibinfo{person}{Brian Yang}, \bibinfo{person}{Huangyuan Su}, \bibinfo{person}{Nikolaos Gkanatsios}, \bibinfo{person}{Tsung-Wei Ke}, \bibinfo{person}{Ayush Jain}, \bibinfo{person}{Jeff Schneider}, {and} \bibinfo{person}{Katerina Fragkiadaki}.} \bibinfo{year}{2024}\natexlab{}.
\newblock \showarticletitle{Diffusion-es: Gradient-free planning with diffusion for autonomous driving and zero-shot instruction following}.
\newblock \bibinfo{journal}{\emph{arXiv preprint arXiv:2402.06559}} (\bibinfo{year}{2024}).
\newblock


\bibitem[Yang et~al\mbox{.}(2023)]%
        {yang2023llm4drive}
\bibfield{author}{\bibinfo{person}{Zhenjie Yang}, \bibinfo{person}{Xiaosong Jia}, \bibinfo{person}{Hongyang Li}, {and} \bibinfo{person}{Junchi Yan}.} \bibinfo{year}{2023}\natexlab{}.
\newblock \showarticletitle{Llm4drive: A survey of large language models for autonomous driving}.
\newblock \bibinfo{journal}{\emph{arXiv preprint arXiv:2311.01043}} (\bibinfo{year}{2023}).
\newblock


\bibitem[Yao et~al\mbox{.}(2024)]%
        {yao2024calmm}
\bibfield{author}{\bibinfo{person}{Ruoyu Yao}, \bibinfo{person}{Yubin Wang}, \bibinfo{person}{Haichao Liu}, \bibinfo{person}{Rui Yang}, \bibinfo{person}{Zengqi Peng}, \bibinfo{person}{Lei Zhu}, {and} \bibinfo{person}{Jun Ma}.} \bibinfo{year}{2024}\natexlab{}.
\newblock \showarticletitle{CALMM-Drive: Confidence-Aware Autonomous Driving with Large Multimodal Model}.
\newblock \bibinfo{journal}{\emph{arXiv preprint arXiv:2412.04209}} (\bibinfo{year}{2024}).
\newblock


\bibitem[Yuan et~al\mbox{.}(2024)]%
        {yuan2024rag}
\bibfield{author}{\bibinfo{person}{Jianhao Yuan}, \bibinfo{person}{Shuyang Sun}, \bibinfo{person}{Daniel Omeiza}, \bibinfo{person}{Bo Zhao}, \bibinfo{person}{Paul Newman}, \bibinfo{person}{Lars Kunze}, {and} \bibinfo{person}{Matthew Gadd}.} \bibinfo{year}{2024}\natexlab{}.
\newblock \showarticletitle{Rag-driver: Generalisable driving explanations with retrieval-augmented in-context learning in multi-modal large language model}.
\newblock \bibinfo{journal}{\emph{arXiv preprint arXiv:2402.10828}} (\bibinfo{year}{2024}).
\newblock


\bibitem[Zeng et~al\mbox{.}(2019)]%
        {zeng2019end}
\bibfield{author}{\bibinfo{person}{Wenyuan Zeng}, \bibinfo{person}{Wenjie Luo}, \bibinfo{person}{Simon Suo}, \bibinfo{person}{Abbas Sadat}, \bibinfo{person}{Bin Yang}, \bibinfo{person}{Sergio Casas}, {and} \bibinfo{person}{Raquel Urtasun}.} \bibinfo{year}{2019}\natexlab{}.
\newblock \showarticletitle{End-to-end interpretable neural motion planner}. In \bibinfo{booktitle}{\emph{Proceedings of the IEEE/CVF conference on computer vision and pattern recognition}}. \bibinfo{pages}{8660--8669}.
\newblock


\bibitem[Zhang et~al\mbox{.}(2020)]%
        {zhang2020autosys}
\bibfield{author}{\bibinfo{person}{Y Zhang}, \bibinfo{person}{S Zhang}, \bibinfo{person}{Y Zhang}, \bibinfo{person}{J Ji}, \bibinfo{person}{Y Duan}, \bibinfo{person}{Y Huang}, \bibinfo{person}{J Peng}, {and} \bibinfo{person}{Y Zahng}.} \bibinfo{year}{2020}\natexlab{}.
\newblock \showarticletitle{Multi-modality fusion perception and computing in autonomous driving}.
\newblock \bibinfo{journal}{\emph{J. Comput. Res. Dev}}  \bibinfo{volume}{57} (\bibinfo{year}{2020}), \bibinfo{pages}{1781--1799}.
\newblock


\bibitem[Zheng et~al\mbox{.}(2024)]%
        {zheng2024planagent}
\bibfield{author}{\bibinfo{person}{Yupeng Zheng}, \bibinfo{person}{Zebin Xing}, \bibinfo{person}{Qichao Zhang}, \bibinfo{person}{Bu Jin}, \bibinfo{person}{Pengfei Li}, \bibinfo{person}{Yuhang Zheng}, \bibinfo{person}{Zhongpu Xia}, \bibinfo{person}{Kun Zhan}, \bibinfo{person}{Xianpeng Lang}, \bibinfo{person}{Yaran Chen}, {et~al\mbox{.}}} \bibinfo{year}{2024}\natexlab{}.
\newblock \showarticletitle{Planagent: A multi-modal large language agent for closed-loop vehicle motion planning}.
\newblock \bibinfo{journal}{\emph{arXiv preprint arXiv:2406.01587}} (\bibinfo{year}{2024}).
\newblock


\bibitem[Zhu et~al\mbox{.}(2020)]%
        {zhu2020deformable}
\bibfield{author}{\bibinfo{person}{Xizhou Zhu}, \bibinfo{person}{Weijie Su}, \bibinfo{person}{Lewei Lu}, \bibinfo{person}{Bin Li}, \bibinfo{person}{Xiaogang Wang}, {and} \bibinfo{person}{Jifeng Dai}.} \bibinfo{year}{2020}\natexlab{}.
\newblock \showarticletitle{Deformable detr: Deformable transformers for end-to-end object detection}.
\newblock \bibinfo{journal}{\emph{arXiv preprint arXiv:2010.04159}} (\bibinfo{year}{2020}).
\newblock


\end{thebibliography}
